\documentclass[twoside,11pt]{article}

\usepackage{jmlr2e}

\usepackage{amsmath,amsfonts}
\usepackage{algorithmic}
\usepackage{array}
\usepackage[caption=false,font=normalsize,labelfont=sf,textfont=sf]{subfig}
\usepackage{textcomp}
\usepackage{stfloats}
\usepackage{url}
\usepackage{verbatim}
\usepackage{graphicx}

\usepackage{mathrsfs}
\usepackage{placeins}
\usepackage[norule]{footmisc}

\newcommand{\clock}{
 {\mathchoice
  {\includegraphics[height=1.6ex]{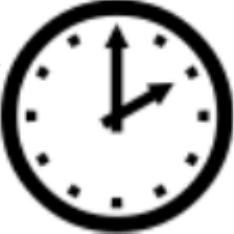}}
  {\includegraphics[height=1.6ex]{clock_symbol}}
  {\includegraphics[height=1.2ex]{clock_symbol}}
  {\includegraphics[height=0.9ex]{clock_symbol}}
 }
}

\newcommand{\cOx}{\text{C$\clock$x}}

\usepackage{lastpage}
\jmlrheading{??}{2023}{1-\pageref{LastPage}}{?/??; Revised ?/??}{?/??}{??-0000}{S\"oren Dittmer, Michael Roberts, Jacobus Preller, AIX COVNET, James H.F. Rudd, John A.D. Aston, and Carola-Bibiane Sch\"onlieb}

\ShortHeadings{Reinterpreting survival analysis in the universal approximator age}{Reinterpreting survival analysis in the universal approximator age}
\firstpageno{1}

\begin{document}
\title{Reinterpreting Survival Analysis in the Universal Approximator Age}

\author{
    \name S\"oren Dittmer \email sd870@cam.ac.uk \\
    \addr Cambridge Image Analysis Group,
    University of Cambridge, UK\\
    Center for Industrial Mathematics,
    University of Bremen, Germany
\AND
    \name Michael Roberts \email mr808@cam.ac.uk \\
    \addr Cambridge Image Analysis Group,
    University of Cambridge, UK\\
\AND
    \name Jacobus Preller \email jacobus.preller1@nhs.net \\
    \addr Department of Medicine,
    University of Cambridge, UK\\
\AND
    \name AIX COVNET \\
    \addr Cambridge Image Analysis Group,
    University of Cambridge, UK\\
\AND
    \name James H.F. Rudd \email jhfr2@cam.ac.uk \\
    \addr Department of Medicine,
    University of Cambridge, UK\\
\AND
    \name John A.D. Aston \email jada2@cam.ac.uk \\
    \addr Statistical Laboratory,
    University of Cambridge, UK\\
\AND
    \name Carola-Bibiane Sch\"onlieb \email cbs31@cam.ac.uk \\
    \addr Cambridge Image Analysis Group,
    University of Cambridge, UK\\
}


\maketitle

\begin{abstract}
Survival analysis is an integral part of the statistical toolbox. However, while most domains of classical statistics have embraced deep learning, survival analysis only recently gained some minor attention from the deep learning community. This recent development is likely in part motivated by the COVID-19 pandemic. We aim to provide the tools needed to fully harness the potential of survival analysis in deep learning. On the one hand, we discuss how survival analysis connects to classification and regression. On the other hand, we provide technical tools. We provide a new loss function, evaluation metrics, and the first universal approximating network that provably produces survival curves without numeric integration. We show that the loss function and model outperform other approaches using a large numerical study.
\end{abstract}

\let\thefootnote\relax\footnotetext{Code available at \url{https://github.com/sdittmer/survival_analysis_sumo_plus_plus}}

\section{Introduction}
Survival analysis (SA) is a well-established branch of statistics~\cite{cox2018analysis, kalbfleisch2011statistical, kleinbaum2012survival, yang2011oasis, han2016oasis}. Nevertheless, it has only recently been encountered in the world of machine learning and, even more recently, deep learning (DL)~\cite{lee2015bflcrm, kong2018flcrm, lee2018deephit, giunchiglia2018rnn, ching2018cox, yang2018spatio}; see~\cite{wang2019machine, sonabend2021theoretical} for an overview on classical machine learning for SA. As the DL community has only just started focusing on SA, many fundamental questions remain unsettled.

This paper aims to provide the tools to make SA as valuable for the DL community as possible through four main contributions. 1. We examine how SA can bridge the well-studied domains of classification and regression. 2. We propose a novel loss function for SA in the continuous-time setting. 3. We present the ﬁrst DL model that provably produces survival curves in continuous time without the need for numeric integration methods and is also a universal approximator. 4. We propose a modified integrated Brier score incorporating classifier metrics for a more nuanced evaluation of highly expressive survival models.

We hope our contributions will enable a more flexible and widespread application of SA in DL. As one can apply SA to predict death, but also the occurrence of any event, we see great potential for deeper integration of the two subjects. For example, there is a significant interest in applying SA to a variety of data modalities, e.g., omics data~\cite{zhang2022survbenchmark} and imaging data~\cite{shahin2022survival}. Particularly since 2020, with its utilization during the COVID-19 pandemic~\cite{wiegand2022development,crooks2020predicting,schwab2021real}, there has been increased interest in the adoption of deep learning for SA. 

We will now describe the structure of the paper and its relation to previous work on the subject.

In Section~\ref{sec:reinterpret}, we examine the relationship between SA, classification, and regression. This connection has been partially examined in prior works such as~\cite{zhong2019survival}, but not yet thoroughly. Other classification-inspired works have adapted binary cross-entropy losses for SA~\cite{lee2018deephit, ren2019deep}. Our approach differs by formulating a loss function for continuous time settings, leveraging a probabilistic perspective.

In Section~\ref{sec:universal_and_exclusive}, we first extend the monotonically increasing MONDE networks~\cite{chilinski2020neural} by improving their ability to model complex survival curves.
Secondly, we modify the SuMo networks~\cite{rindt2022survival} to not only enable universal approximation but to guarantee that any output is a survival curve. This modification also brings the SuMo network's formulation closer to one of classical survival models.

In Section~\ref{sec:cox_main_section}, we implement a neural network version of the well-known Cox model and its time-dependent variant using methods from Section~\ref{sec:universal_and_exclusive}. This synthesis enables flexible training setups for Cox models and, especially for the time-dependent model, combines the excellent approximation properties of neural nets with the interpretability of Cox models. The papers~\cite{katzman2018deepsurv, shahin2022survival, kvamme2019time} are closest to this part of our work. But, in contrast, these papers replace the Cox model's linear regression with a neural network, not their baseline function.

In Section~\ref{sec:numerics}, we provide a large-scale comparison of different survival models trained with different loss functions. We use seven datasets, one includes image data, and two are originally regression datasets, demonstrating the applicability of survival models to regression tasks. We choose these datasets for their diversity and to analyze our methodology on real-life datasets.
Here we also introduce the modified integrated Brier scores~\cite{graf1999assessment}. Along with the established concordance index, they open up a rich pool of well-known quality measures, enabling a nuanced exploration of different aspects of a given model.

\section{Reinterpreting survival analysis}
\label{sec:reinterpret}
We now discuss how SA can be a portal between classification and regression.

To set the stage, we introduce some key notation. We describe a potentially right-censored survival data sample via
\begin{equation}
    \label{eq:sample}
    (x, e, T) \in \mathbb{R}^n \times \{0, 1\} \times \mathbb{R}_{>0},
\end{equation}
where $x\in\mathbb{R}^n$ denotes input features, $e\in\{0,1\}$ indicates if the event was observed (1) or not (0), and $T$ is the time of the event or the time of right-censoring. We set $T>0$ as one usually assumes that no events occur for $T\le0$.

The goal of SA is to use a set of such samples to create a model that uses the features $x$ to predict the probability of a singular event, such as death, occurring after any given time $t\ge0$. One assumes that the event has not occurred at $t=0$ and is irreversible. Hence, one aims to produce a non-negative, monotonically decreasing curve being $1$ at time $t=0$. Throughout this paper, we will use the terms ``alive'' and ``dead'' to refer to the state before and after the singular event.

\subsection{Survival analysis is classification (with infinitely many classifiers)}
\label{sec:sa_as_classification}
We will pose SA in terms of classification. We begin by recalling the general definition of a survival curve:
\begin{equation}
    \label{eq:S}
    S(t|x): \mathbb{R}_{\ge0}\times\mathbb{R}^n \mapsto \mathbb{P}(\mbox{alive}@t |x) \in [0, 1]
\end{equation}
That is, $S(t|x)$ gives us the probability that the event has not occurred before or at time $t$, given the features $x$.

An alternative way of reading this definition is that $S$ is equivalent to a set of probabilistic binary classifiers
\begin{equation}
    \{S(t|\cdot) = C_t:\mathbb{R}^n\to[0,1]:t\in\mathbb{R}_{\ge0}\},
\end{equation}
each $C_t(x)$ being the probability of surviving the interval $[0, t]$ given the features $x$. Phrased differently, \textit{a survival model is simply an infinite collection of probabilistic classifiers indexed by time}.

Using the classifier perspective, we will derive a new loss function for survival curves. Considering the ideal scenario of having unlimited clean data and a highly capable probabilistic classifier, we can train one classifier for each point in time. This transforms the problem into a prediction of Bernoulli variables, and thus each classifier can be trained using the binary cross entropy (BCE) loss.

As a reminder, the standard BCE loss for a classifier $C:\mathbb{R}^n\to[0,1]$ and a sample $(x, y)\in\mathbb{R}^n\times\{0,1\}$ is defined as
\begin{equation}
    \label{eq:standard_bce}
    -y\log C(x) - (1-y)\log\left(1 - C(x)\right).
\end{equation}

Returning to the SA setting, if we have a censored sample, $(x, 0, T)$, we can use it to train all classifiers before the censoring,
\begin{equation}
    \label{eq:set_early_classifiers}
    \{C_t:\mathbb{R}^n\to[0,1]:t\in[0, T)\},
\end{equation}
via the BCE part
\begin{equation}
    \label{eq:loss_early_classifiers}
    -\log C_t(x),
\end{equation}
i.e., for censored samples, we want to find classifiers that predict $0$ (``alive'') for times before the censoring.

Otherwise, if we have an uncensored sample, $(x, 1, T)$, we can use the loss in~\eqref{eq:loss_early_classifiers} for the early (``premortem'') classifiers given by~\eqref{eq:set_early_classifiers} and train the later (``postmortem'') classifiers given by
\begin{equation}
    \label{eq:set_later_classifiers}
    \{C_t:\mathbb{R}^n\to[0,1]:t\ge T\}
\end{equation}
via the BCE part
\begin{equation}
    \label{eq:loss_later_classifiers}
    -\log(1 - C_t(x)).
\end{equation}

We can formulate a joint loss for all classifiers by choosing two arbitrary random variables, $\mathcal{T}_-(e, T)$ and $\mathcal{T}_+(e, T)$, with an everywhere-supported density over $[0, T]$ and $[T, \infty)$ respectively.
We define this loss by joining~\eqref{eq:loss_early_classifiers} and~\eqref{eq:loss_later_classifiers} to
\begin{align}
\label{eq:bce_loss}
\begin{split}
    L_\text{BCE} = -[
                   & \mathbb{E}_{t_-\sim \mathcal{T}_-}\log S(t_-|x) \\
    + e \, \cdot\, & \mathbb{E}_{t_+\sim \mathcal{T}_+}\log \left(1 - S(t_+|x)\right)
    ].
\end{split}
\end{align}
The loss trains a separate classifier $S(t|\cdot)=C_t(x)$ for every point in time $t\in[0,\infty)$, as both $\mathcal{T}_\pm$ are supported everywhere. We can consider the loss as a weighted integral over the BCE loss. As a result, it inherits many desirable properties of the BCE loss, such as being a strictly proper scoring rule~\cite{du2021beyond}.

In practice, assuming an infinite amount of clean data is unrealistic. However, we have two advantages when dealing with commonly found un- and right-censored data. Firstly, earlier classifiers can use most data points (as little censoring has occurred yet). Secondly, as survival curves decrease over time, later classifiers' predictions are upper bounded by earlier classifiers. To take advantage of this, we need to ensure the monotonicity of our predicted survival curves, which we will discuss in Section~\ref{sec:universal_and_exclusive}.

If the monotonicity holds, we also do not require that the density of $\mathcal{T}_\pm$ is supported everywhere, as later classifiers provide lower bounds for earlier ones. In practice, we therefore do the following: We sample times from a Gaussian centered at $T$ for uncensored samples and only from the left side of that Gaussian for censored samples. We ensure no negative time samples via projection with ReLU. The idea is that while one trains for all $t$, the Gaussian focuses the training effort on values around $T$; here, the variance $\sigma^2$ is a hyperparameter that controls the focus. Other choices are plausible, but we leave this to future work. Formally, we can represent this as
\begin{equation}
    \label{eq:T_minus}
    \mathcal{T}_-(e, T) = 
    \begin{cases}
        \delta(T), & \text{if $e=0$,}\\
        \mbox{ReLU}_\#\mathcal{HN}_-(T, \sigma^2), & \text{otherwise},
  \end{cases}
\end{equation}
and
\begin{equation}
    \label{eq:T_plus}
    \mathcal{T}_+(e, T) = \mathcal{HN}_+(T, \sigma^2),
\end{equation}
where $\delta$ denotes the delta-distribution, $\mathcal{HN}_\pm$ the left- and right-sided half-normal distribution, and $\mbox{ReLU}_\#$ the pushforward operator of the projection into the non-negative real numbers. 

During training, we approximate the expectations in~\eqref{eq:bce_loss} by sampling only one $t$. If $e=0$ we sample $t\sim\mathcal{T}_-$, if $e=1$, we sample with a 50\% probability from $\mathcal{T}_-$ otherwise from $\mathcal{T}_+$.

As a comparison, we use the SuMo loss~\cite{rindt2022survival}
\begin{equation}
    L_\text{SuMo} = -\left[e \log f(t|x) + (1 - e) \log S(t|x)\right],
\end{equation}
where
\begin{equation}
    f(t|x) = - \partial_t S(t|x).
\end{equation}
Both losses can be seen as log-likelihoods and are similar in their $e=1$ part but differ in their $e=0$ part. This is because SuMo's likelihood is conditioned on the event variable $e$, while the BCE's is not.

\subsection{Survival analysis is regression (with uncertainty estimation)}\label{sec:sa_as_regression}
We will now discuss how we can use a survival model to predict the full posterior of $\mathbb{R}^n\to\mathbb{R}_{\ge0}$ regression problems. The idea is similar to the motivation of the paper~\cite{chilinski2020neural}. Using~\eqref{eq:S}, we can express the lifetime distribution function (probability of the event has not occurred yet) as 
\begin{equation}
    F(t|x) = \mathbb{P}(\mbox{dead}@t|x) = 1 - S(t|x).
\end{equation}
Consequently, we can interpret $f(t|x)$ as an event density.

We now want to make a simple but powerful observation. We can treat any dataset with non-negative scalar labels as an uncensored survival dataset. Consequently, the event density $f$ of a survival model that fits the dataset well (see Section~\ref{sec:universal_and_exclusive}) is the full posterior of the original regression problem.

Note that this setup also allows one to train regression models on samples for which one only knows a lower bound on the label.

\subsection{If in doubt: survival analysis}
\begin{figure}
    \centering
    \includegraphics[width=4cm]{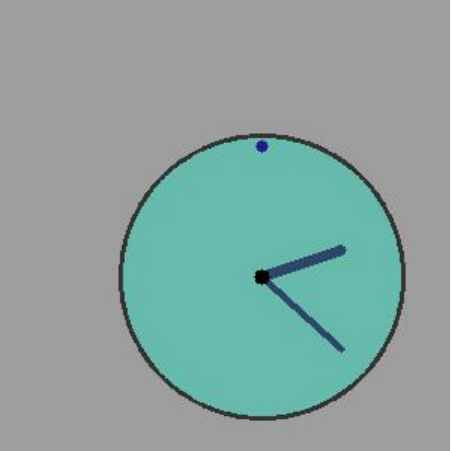} \includegraphics[width=4cm]{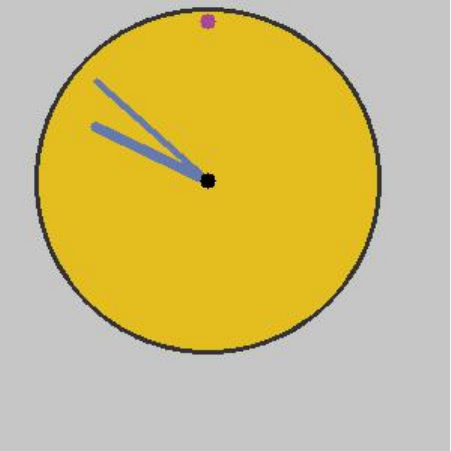}
    \caption{Two samples from the Clocks dataset~\cite{clocks}.}
    \label{fig:clock}
    \vspace{-4mm}
\end{figure}

SA can be viewed from both a classification and regression perspective. Understanding this duality is beneficial when selecting a model or interpreting results. Any one-dimensional regression problem or binary classification involving a threshold can benefit from this perspective by rephrasing the question one asks. For example, the clocks dataset~\cite{clocks} provides images of analog clocks with the displayed times as labels; see Figure~\ref{fig:clock} for two samples. We can interpret the dataset as asking: ``what time does the clock show?''; this is a regression problem. Alternatively, we could ask for any point in time: ``is it earlier than the time the clock shows?''; this subtle shift in question turns the regression into a classification problem. Training a survival model instead of a regression or classification model can answer both questions and many others.

The ability of survival models to answer various questions at inference time is a strong advantage (particularly those not considered during training). Another benefit of framing a problem as a survival problem, where possible, is that the training process receives more information per label. While classification uses binary information about which side of a threshold the value is on, SA uses the value itself.
\section{A universal and exclusive survival curve approximator}
\label{sec:universal_and_exclusive}
We will now introduce MONDE+, a version of MONDE~\cite{chilinski2020neural} more capable of modeling complex survival curves, and SuMo++, an improved version of SuMo~\cite{rindt2022survival} that now guarantees $S(0|x)=1$ and uses MONDE+. This makes the model not only a universal but also an exclusive approximator of survival curves, by which we mean that the model is capable of approximating any survival curve and incapable of producing anything but a survival curve. For a review of MONDE and SuMo, see appendix, Section~\ref{sec:monde_sumo}. We also refer to~\cite{saul2016gaussian} for universal approximators derived via Gaussian processes and~\cite{danks2022derivative} for ones derived via numeric integration of neural networks.

We start by introducing the notation for the cumulative hazard function, $\Lambda(t|x)$, defined via
\begin{equation}
\label{eq:chf_formulation}
S(t|x) = \exp\left[ -\Lambda(t|x)\right].
\end{equation}
One can interpret $\Lambda$ as the total accumulated risk. Since $\Lambda$ determines the model, building a good survival model is equivalent to building a good model for $\Lambda$. The model should be flexible enough to approximate any cumulative hazard. To guarantee survival curves, we have to ensure that $\Lambda$ is monotonically increasing and $\Lambda(0|\cdot)=0$.

\subsection{MONDE+}
As the main building block for such $\Lambda$ we now define the MONDE+ network 
\begin{equation}
    M_+:\mathbb{R} \times \mathbb{R}^n \ni (t, x) \mapsto M_+(t,x) \in \mathbb{R}.
\end{equation}
It is monotonically increasing in its first argument and we define it via layers
\begin{equation}
    \label{eq:monde+}
    z_{k+1}(t, z_k, z_0) = H_kz_k + \sigma_k(\tilde z_k(t, z_k) + B_kz_k + L_kz_0)
\end{equation}
with
\begin{equation}
    \tilde z_k(t, z_k) = A_k\left(\phi_k(a_kt + b_k) \circ \psi_k(G_kz_k)\right)
\end{equation}
where $a_k$ and $b_k$ are vectors, all capitalized letters represent affine maps, and $\circ$ denotes the Hadamard product. $\sigma_k, \phi_k,$ and $\psi_k$ are monotonically increasing functions with non-negative $\phi_k$ and $\psi_k$. In practice, we use $\phi_k=\psi_k=\mbox{softplus}$ and $\sigma_k=\tanh$, in the last layer we set $\sigma_K=\mbox{id}$. Similar to MONDE, we constrain the weight matrices of $A_k, B_k, G_k,$ and $H_k$ and the vector $a_k$ to be pointwise non-negative. Refer to Section~\ref{sec:initalization} of the appendix for details on the initialization.

We recall that a MONDE network  $M:\mathbb{R}^{n+1}\to\mathbb{R}$ consists of layers
\begin{equation}
    \label{eq:monde}
    z_{k+1}(z_k) = \sigma_k(B_kz_k),
\end{equation}
i.e., they cannot differentiate inputs $t, x$, have no residual connections~\cite{he2016deep}, and are a strict subset of the MONDE+ layers.

Unlike MONDE, MONDE+ is monotone in $t$ but has no unnecessary monotonicity constraint in $x$, see the following lemma.
\begin{lemma}
    \label{lemma:monde_plus}
    Let $M_+:\mathbb{R} \times \mathbb{R}^n \to \mathbb{R}$ be a MONDE+ network, i.e., a concatenation of layers defined by~\eqref{eq:monde+}. Then the output of each layer of $M_+$ is pointwise monotonically increasing in $t$.
\end{lemma}

For the proof, see Section~\ref{sec:monde_proof} of the appendix. Each MONDE+ layer contains a MONDE layer as a subset of its operations. As the universal approximation properties of the MONDE network are already established~\cite{lang2005monotonic}, this also makes MONDE+ a universal approximator. As we will show in Section~\ref{sec:numerics}, the MONDE+ component within SuMo++ can successfully model the survival curves for a broad range of applications.

\subsection{SuMo++}
As previously stated, the SuMo++ network aims to improve SuMo's ability to produce accurate survival curves while maintaining its theoretical guarantees of universal approximation.

We define the SuMo++ network as
\begin{equation}
    \mathscr{S}_{++}(t, x) = \exp\left( -\left[M_+(t, q) - M_+(0, q)\right]\right),
\end{equation}
i.e., we set the cumulative hazard function to
\begin{equation}
    \Lambda(t|q) = M_+(t, q) - M_+(0, q)
\end{equation}
with $q = Q(x)\in\mathbb{R}^k$ being some feature extracting network. The exponent in SuMo++ is monotonically decreasing in $t$ and $0$ for $t=0$. Therefore, SuMo++ is not only monotonically decreasing with values in $[0, 1]$, but also guarantees $\mathscr{S}_{++}(0, x)=1$.

In contrast, the SuMo network is defined via a MONDE network as
\begin{equation}
    \mathscr{S}(t, x) = 1 - \mbox{sigmoid}\left(M([t, Q(x)])\right),
\end{equation}
it can, at best, achieve $\mathscr{S}(0, x)=1$ up to some error $\epsilon > 0$ since $\mathscr{S}(t, x) < 1$.

Note that alternatively, one could define SuMo++ via $\mathscr{S}_{++}(t, x) = \frac{1}{1 +\left[M_+(t, q) - M_+(0, q)\right]}.$ However, we did not see any difference in performance, but the former formulation leaves the model and its interpretability closer to classical formulation in~\eqref{eq:chf_formulation}.

We formulated SuMo++ in terms of MONDE+, but we could also do so via MONDE by splitting its input into $z=(t, x)$. For simplicity, we will refer to the SuMo++ approach using MONDE as SuMo+.

Now, as we have established a universal and exclusive approximator of survival curves, we will discuss how SuMo++ is also backward compatible, i.e., how we can use it and MONDE+ to parameterize and treat Cox models as neural networks.
\section{Bringing (time-dependent) Cox models into the deep learning age}
\label{sec:cox_main_section}
The Cox model, introduced in 1972~\cite{cox1972regression}, is arguably the most popular survival model. It uses a baseline cumulative hazard function $\Lambda_0: \mathbb{R}_{\ge0}\ni t \mapsto \Lambda_0(t) \in \mathbb{R}_{\ge0}$, that depends only on time, and a linear regression to modulate the hazard up and down based on the input features $x$.
Formally we write
\begin{equation}
    \mathscr{S}_\text{Cox}(t, x) = \exp\left( -\alpha(x)\Lambda_0(t)\right),
\end{equation}
where
\begin{equation}
    \label{eq:scalar_product_cox}
    \alpha(x) = \exp \left(\langle a, x\rangle\right)
\end{equation}
with the coefficient $a\in\mathbb{R}^n$. This approach seems to strike an outstanding balance between expressiveness and interpretability.

We now describe what we believe is the first implementation of the Cox (and later time-dependent Cox) model that directly parameterizes $ \Lambda_0 $ via a neural network. Note, for the Cox model, \cite{danks2022derivative} implements $\Lambda_0$ indirectly via a neural network but requires the numeric integration of it.

\subsection{A neural network parameterized Cox model}
\label{sec:cox_simple}
We will now introduce a neural network-based implementation of the classical Cox model. While we will use MONDE+ and SuMo++, the following formulations also hold for MONDE and SuMo+ but not SuMo.

The function $\Lambda_0$ is monotonically increasing, with $0$ as a fixed point, and can be parameterized in different ways, e.g., by splines~\cite{efron1988logistic, royston2002flexible} or in a non-parametric fashion~\cite{crowley1984statistical}. We will do so using neural networks.

Using the results in Section~\ref{sec:universal_and_exclusive}, we can parameterize a Cox model by setting
\begin{equation}
    \label{eq:lambda0}
    \Lambda_0(t) = M_+(t, 0) - M_+(0, 0)
\end{equation}
and obtain
\begin{align}
\begin{split}
    \mathscr{S}_\text{Cox}(t, x)
    = \exp\left( -\alpha(x)\Lambda_0(t)\right) 
    = \mathscr{S}_{++}(t, 0)^{\alpha(x)}.
\end{split}
\end{align}
We want to emphasize that in contrast to the DeepSurv model (also known as DeepHit) independently introduced by~\cite{katzman2018deepsurv},~\cite{lee2018deephit} and~\cite{shahin2022survival}, we propose a parameterization for $\Lambda_0$. This makes our approach not a modification, but an implementation, of the Cox model and, therefore, equally interpretable. Still, we enable training and evaluation within DL frameworks using automatic differentiation~\cite{paszke2017automatic}. This enables not only easy training of Cox models using different loss functions, but also straightforward computation of the event density, $f=-\partial_t S$, and hazard function $\lambda(t|x) = \partial_t\Lambda(t|x)$.

\subsection{A neural network parameterized time-dependent Cox model}
\label{sec:cox_time_dependent}
We will now show how we can parameterize Cox models with time-dependent coefficients~\cite{fisher1999time} using neural networks. These models are of the form
\begin{equation}
    \label{eq:cOx}
    \mathscr{S}_\cOx(t, x) = \exp\left( -\alpha(t, x)\Lambda_0(t)\right),
\end{equation}
where one usually sets $\alpha(t, x) = \exp\left[ \langle \omega(t), x\rangle\right]$ with $\omega: \mathbb{R}_{\ge0} \to \mathbb{R}^n$.

Again, as in Section~\ref{sec:cox_simple}, we can parameterize $\Lambda_0(t)$ via~\eqref{eq:lambda0}. For $\alpha$, we propose the parametrization
\begin{align}
\label{eq:cOx_alpha}
    \alpha(t, x) = \exp\left( \langle \omega(t, x), |x - o|\rangle\right)
\end{align}
where $|\cdot|$ is the entrywise absolute value and $o\in\mathbb{R}^n$ a learnable offset parameter. We use two MONDE+ networks $M_+^\pm$ to define
$\omega:\mathbb{R}_{\ge0}\times \mathbb{R}^n\to \mathbb{R}^n$
entrywise as
\begin{equation}
    \omega_i(t, x) =
    \begin{cases}
        \beta^-_i(t), & \text{if $x_i<o_i$,}\\
        \beta^+_i(t), & \text{otherwise},
  \end{cases}
\end{equation}
where we set $\beta^\pm:\mathbb{R}_{\ge0} \to \mathbb{R}^n$ to
\begin{equation}
    \beta^-(t) = M_+^-(t, 0) - M_+^-(0, 0) + M_+^+(0, 0)
\end{equation}
and
\begin{equation}
    \beta^+(t) = M_+^+(t, 0).
\end{equation}

Unlike in Section~\ref{sec:cox_simple}, this is not just a reparametrization but a slight modification guaranteeing $\mathscr{S}_\cOx(t, x)$ to decrease monotonically in $t$, see Lemma~\ref{lemma:cOx}. Still, the coefficients $\omega$ are highly interpretable, as they separate the effects of $t$ from $x$. Essentially, $\omega_i$ provides a time-dependent weight for $x_i$ that is independent of $x$ except for the threshold $o_i$, which marks the ``least dangerous'' value $x_i$ can take.
\begin{lemma}
    \label{lemma:cOx}
    For the model $\mathscr{S}_\cOx(t, x)$, defined by~\eqref{eq:cOx} and \eqref{eq:cOx_alpha}, it holds $\forall x \in\mathbb{R}^n$ that:
    \begin{enumerate}
        \item $\mathscr{S}_\cOx(t, x)\in[0,1]$ for all $t\in\mathbb{R}_{\ge0}.$
        \item We have a maximum $\mathscr{S}_\cOx(0, x)=1$.
        \item $\mathscr{S}_\cOx(t, x)$ is monotonically decreasing in $t$.
    \end{enumerate}
    Further, if the $M_+^\pm$ are continuous in $t$, we have:
    \begin{enumerate}
        \item The $\omega_i$ are continuous in $t$.
        \item For any $x$, the $\omega_i(\cdot, x)$ have their minima at $\omega_i(0, x)=\beta_i^+(0).$
        \item $\alpha$ is continuous in $(t, x)$.
    \end{enumerate}
\end{lemma}
See Section~\ref{sec:proof_cOx} of the appendix for the proof.

\section{Numerical experiments}
\label{sec:numerics}
We will now present the results of our numerical experiments. We will first compare the performance of MONDE and MONDE+ and then perform an extensive comparison of different survival models using several survival and regression datasets.
\subsection{MONDE vs. MONDE+: A toy example}
\label{sec:toy_example}
We now compare the performance of the SuMo, SuMo+, and SuMo++ networks -- i.e., MONDE without and with the initial condition enforced and MONDE+ with the initial condition enforced.

To showcase the limitations of the SuMo and SuMo+ models in recreating survival curves, we designed an experiment using six samples, each containing 32 features, each in the range $[-1,1]$. We assigned survival probabilities for each sample at 2-7 points over the time interval $[0,1]$ in a way that tests the models' ability to capture complex distributions, such as sharp declines followed by plateaus and vice versa.

\begin{figure}
    \centering
    \includegraphics[width=.56\textwidth, trim={10mm 5mm 15mm 0mm}, clip]{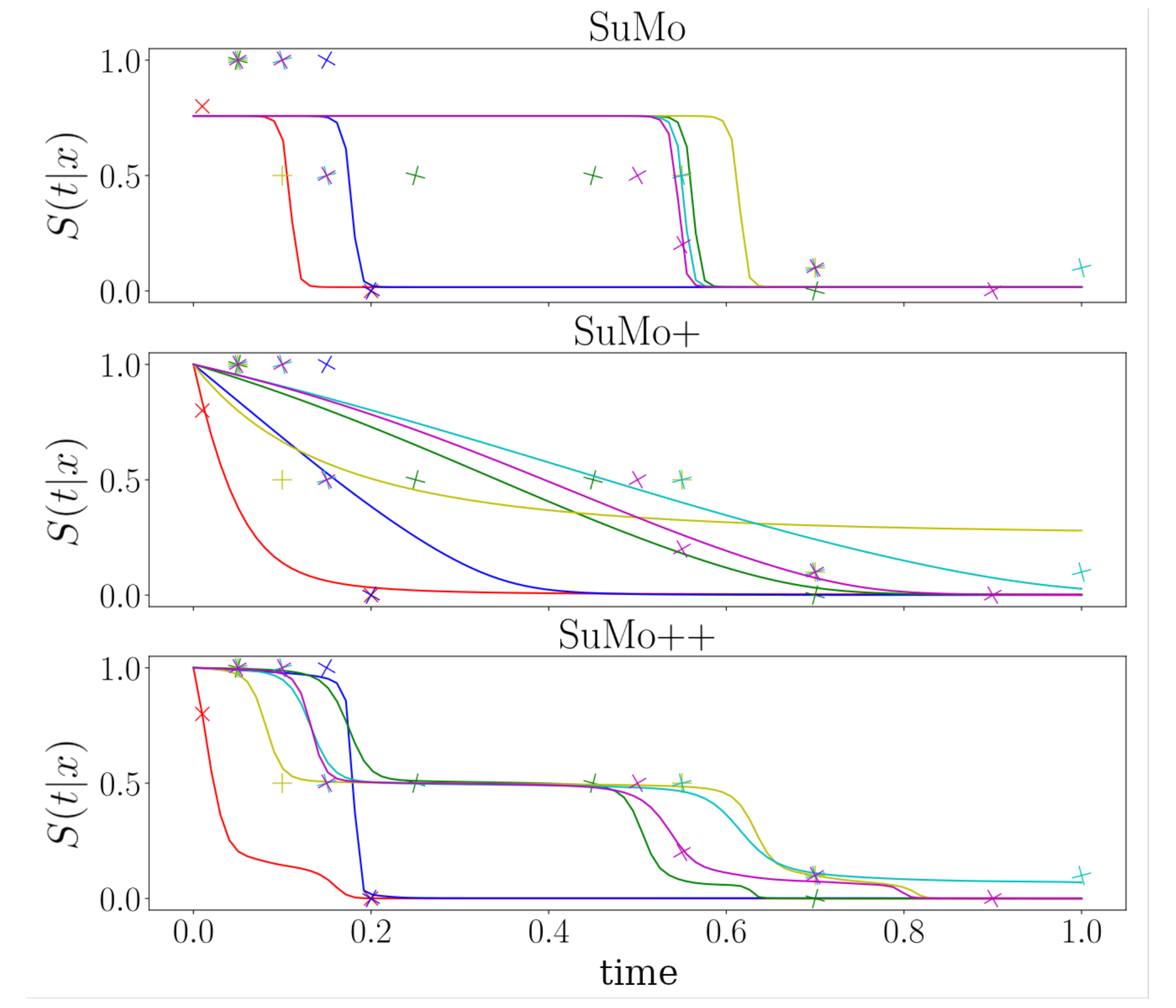}
    \caption{Points and their corresponding learned curves are color-matched. To distinguish overlapping points, we use crosses with varying rotations. Each of the three plots is the result of one training. The plots correspond to the losses of $2.5$, $2.6$, and $1.7$ (top to bottom). The top plot shows that the SuMo network can produce curves that do not fulfill survival curve's initial condition $S(0|x) = 1$.}
    \label{fig:toy_curves}
\end{figure}

\begin{figure}
    \centering
    \includegraphics[width=.56\textwidth, trim={30mm 60mm 35mm 55mm}, clip]{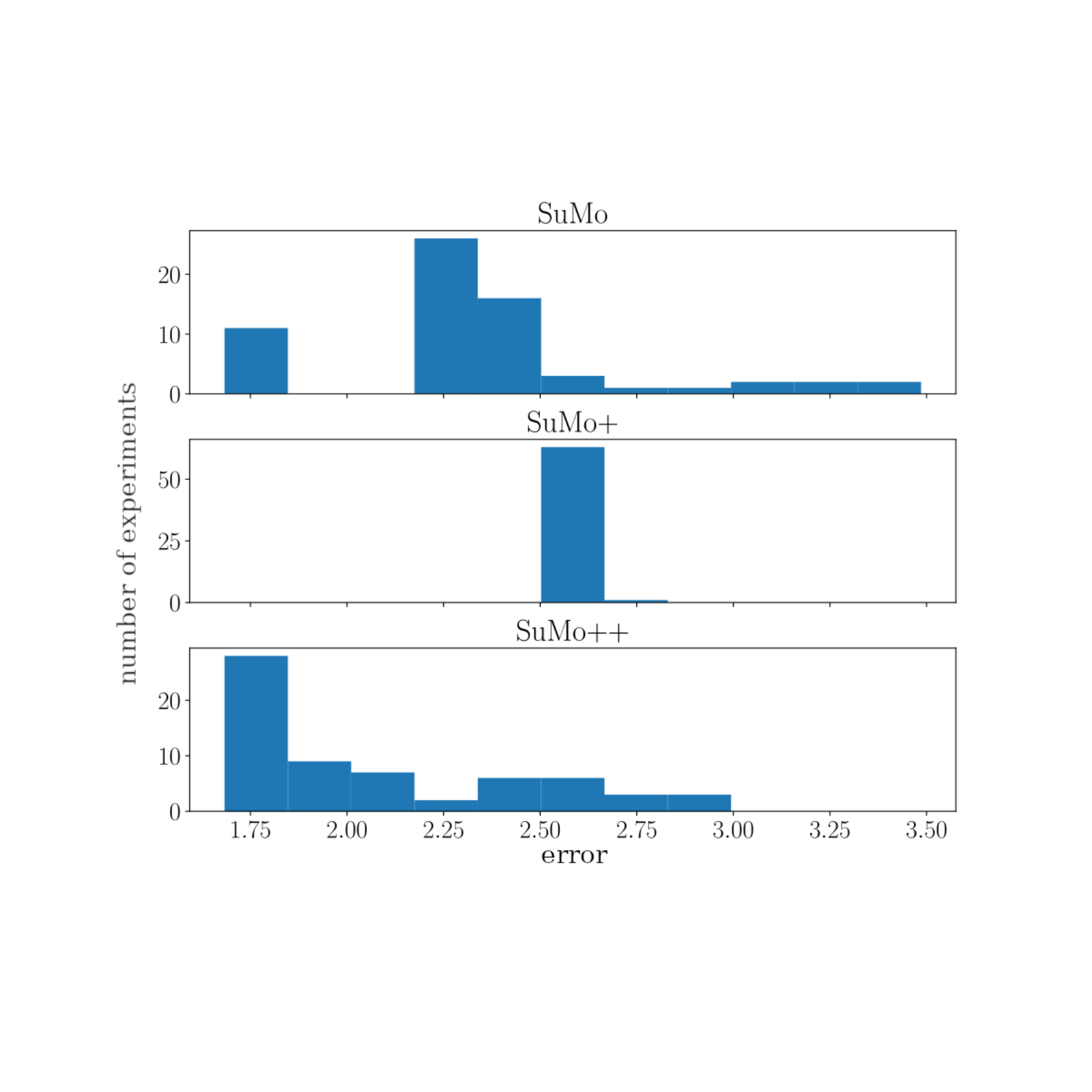}
    \caption{BCE loss histograms of the different networks. Each contains 64 training runs with 512 training steps each. To see the results after convergence (2048 steps), see Figure~\ref{fig:hists_2048} in the appendix. Figure~\ref{fig:hists_2048} shows that both SuMo+ and SuMo++ benefit from further training steps -- SuMo+ almost improves to the level of SuMo++ -- while SuMo shows no significant improvement.}
    \label{fig:hists}
\end{figure}

We trained each network for 512 Adam~\cite{kingma2014adam} iterations with a step size of $10^{-3}$ using the standard binary cross-entropy loss. Figure~\ref{fig:toy_curves} compares the three models over the six samples. To show the variability of the outcome, we also provide histograms of multiple runs of these experiments in Figure~\ref{fig:hists}. The histograms show that it can be difficult for the SuMo+ model to fit the six curves, while the SuMo and SuMo++ can produce better fits.

During these and all following experiments, we used MONDE+ with five hidden layers of width 32. We also use MONDE with the same structure, but after the input layer, we change the width to $98$ to give both networks approximately $32,000$ trainable parameters.

\subsection{Classifier metrics}
\label{sec:classifier_metrics}
We will now discuss how we will evaluate the following experiments. As discussed in~\cite{rindt2022survival}, the classical scoring rules for survival models do not adequately reflect the performance of survival models. Classical scoring rules for survival models, such as time- or hazard-based concordance scores for traditionally trained Cox models, are only effective in their intended context. However, they may not accurately reflect performance in a general setting defined by~\eqref{eq:S}. For example, a model that incorrectly predicts all patients dying within milliseconds of a clinical trial may still have a perfect concordance score, as the score only considers the correct order of events and not their absolute time. This does not make time concordance useless for these models but insufficient. Ideally, one would have multiple scoring rules to choose from to judge the different aspects of a model that are relevant to a given situation.

We propose to use the general integrated Brier score (IBS). However, instead of using the mean squared as the integrated scoring rule, we propose to use general classifier scoring rules, e.g., accuracy. Interestingly this fits squarely into the original general definition of the IBS~\cite{graf1999assessment}, but we have not seen it applied this way.

As a survival model is an infinite number of probabilistic classifiers indexed by time, we can evaluate the survival model's quality from different aspects; simply by choosing from the rich pool of classifier quality measures. E.g., one could use the $F_2$-score if one considers a clinical setting where an overly optimistic prognosis may deny patients timely access to escalation.

Note that we will use threshold-free versions of all classifier scoring rules. For example, if $l\in\{0,1\}^K$ is a binary vector of length $K\in\mathbb{N}$ containing labels and $p\in[0,1]^K$ a vector containing corresponding probabilistic predictions, we define the true-positives to be their scalar product, i.e., $\sum_{k=1}^K p_kl_k$.

The IBS integrates its particular scoring rule over a given time interval. We will use the interval $[0, T_{max}]$ where we choose $T_{max}$ for each dataset based on the $90^{th}$ percentile of times provided by the dataset.

The mean squared error is arguably the most popular scoring rule in regression. Thus the standard IBS neatly connects to the regression point of view we laid out in Section~\ref{sec:sa_as_regression}, while the classifier scoring rules proposed by us connect to Section~\ref{sec:sa_as_classification}.

\subsection{Datasets}
\label{sec:datasets}
We will now provide a brief qualitative overview of the datasets we use in Section~\ref{sec:results_and_discussion}. See Table~\ref{tab:dataset_overview} for a high-level overview. Each dataset consists of samples given by triples as in~\eqref{eq:sample} i.e., $(x, e, T)$. We convert the Clocks and California regression datasets into this format by assigning $T$ the label's value and $e=1$.
\begin{table}
    \centering

    \begin{tabular}{|c||c|c|c|}
        \hline
        Name & \#Samples & \#Features & Test set \\
        \hline
        GBSG2 & 686 & 8 & 25\% \\
        Recur & 1,296 & 4 & 25\% \\
        NKI & 272 & 9 & 25\% \\
        Lymph & 686 & 8 & 25\% \\
        COVID-19 & 1,489 & 16 & 25\% \\
        Clocks & 10,000 & 6,912 & 2.5\% \\
        California & 20,640 & 8 & 2.5\% \\
        \hline
    \end{tabular}

    \caption{An overview of the datasets as we used them. Up to rounding errors, we split the data such that the validation and test have the same size, i.e., if the test set contains 25\% of the sample, the validation contains 25\%, and the training contains 50\% of the samples. For the references see, GBSG2 \cite{sauerbrei1999building}, Recur \cite{lemeshow2011applied}, NKI \cite{nicolau2011topology}, Lymph \cite{schmoor2000role}, Clocks \cite{clocks}, and California \cite{pace1997sparse}.}
    \label{tab:dataset_overview}
\end{table}

The COVID-19 dataset is a private dataset based on COVID-19 patient data from Addenbrooke's Hospital, Cambridge, UK. The seven datasets showcase diverse survival curve dynamics. For example, the dynamics of oncology treatment outcomes and that of critically ill patients differ greatly. The Clocks dataset, with deterministic outcomes, highlights sudden changes in survival curves.

As some of the datasets have relatively few samples, the random seed used to split the data into training, validation, and test sets can significantly impact the evaluation of a model. To mitigate this effect, we ran 1000 splits and selected the seed that minimizes the difference between the three Kaplan-Meier estimators for the training, validation, and test sets. Since Kaplan-Meier estimates the expected unconditioned survival curve for each set, this decreases the chance of having vastly different training, validation, and test sets.

We normalize all input features to have a $0$ mean and standard deviation of $1$. We normalize the times within the data by dividing through by $T_{max}$ as defined in Section~\ref{sec:classifier_metrics}. For more details on the datasets, see Section~\ref{sec:datasets_appendix} in the appendix.

\subsection{Results and discussion}
\label{sec:results_and_discussion}

For the Clocks dataset, containing images, we apply the methods amenable to convolutions to extract the features for the model, namely CoxDeepNN, SuMo, SuMo+, and SuMo++, as well as Kaplan-Meier. For all other datasets, which contain tabular data, we were able to train all models discussed earlier.
See Section~\ref{sec:training_details} in the appendix for training details. We used the standard and classifier metric versions of the IBS discussed in Section~\ref{sec:classifier_metrics} to evaluate the models. We use the classifier metrics accuracy, area under the precision-recall curve (AUPRC), area under the receiver operating characteristic curve (AUROC), balanced accuracy, the $F_\beta$-score for $\beta\in\{0.5, 1, 2\}$, precision, sensitivity, specificity, and the Youden's index~\cite{youden1950index}. In addition to the IBS, we also used the established concordance index based on the restricted mean survival time, restricted by the dataset's $T_{max}$. See Table~\ref{tab:stats_Concordance} and Table~\ref{tab:stats_iibs} in the appendix for the concordance index and the standard IBS. We report the mean of all scores in Table~\ref{tab:stats_Mean}. Note that the CoxDeepNN model approach is well known from other papers~\cite{katzman2018deepsurv, kvamme2019time, shahin2022survival}, commonly referred to as DeepHit or DeepSurv.
\begin{table*}[ht]
\centering
\begin{tabular}{l||rrrrr|rr}
Model Loss-function & COVID-19 & NKI & BGSG2 & Recur & Lymph & Clocks & California \\
\hline \hline
Kaplan-Meier & 0.40 & 0.43 & 0.45 & 0.52 & 0.42 & 0.51 & 0.52 \\
Weibull & 0.67 & 0.52 & 0.51 & 0.75 & 0.47 & $\emptyset$ & 0.68 \\
Log-logistic & 0.67 & 0.53 & 0.51 & 0.76 & 0.47 & $\emptyset$ & 0.71 \\
Log-normal & 0.67 & 0.53 & 0.51 & 0.76 & 0.48 & $\emptyset$ & 0.70 \\
Cox-piecewise & 0.67 & 0.52 & 0.51 & 0.75 & 0.47 & $\emptyset$ & 0.67 \\
Cox-spline & 0.67 & 0.52 & 0.51 & 0.75 & 0.47 & $\emptyset$ & 0.67 \\
\hline
CoxNN SuMo & 0.70 & 0.54 & 0.50 & 0.77 & 0.49 & $\emptyset$ & 0.68 \\
CoxNN BCE & 0.72 & 0.54 & 0.51 & 0.73 & 0.48 & $\emptyset$ & 0.64 \\
\cOx NN SuMo & 0.73 & 0.56 & 0.51 & 0.79 & 0.50 & $\emptyset$ & 0.65 \\
\cOx NN BCE & 0.70 & 0.56 & 0.51 & 0.79 & 0.48 & $\emptyset$ & 0.62 \\
CoxDeepNN SuMo & 0.73 & 0.53 & \textbf{0.55} & 0.79 & 0.50 & 0.69 & 0.69 \\
CoxDeepNN BCE & 0.72 & 0.58 & 0.49 & 0.78 & 0.49 & 0.83 & 0.65 \\
\hline
SuMo SuMo & 0.73 & 0.55 & 0.49 & 0.80 & 0.50 & \textbf{0.97} & 0.80 \\
SuMo BCE & 0.74 & 0.62 & 0.54 & \textbf{0.81} & 0.47 & 0.91 & 0.77 \\
SuMo+ SuMo & 0.72 & 0.50 & 0.50 & 0.79 & 0.49 & \textbf{0.97} & \textbf{0.81} \\
SuMo+ BCE & \textbf{0.76} & 0.59 & 0.54 & 0.79 & 0.48 & 0.91 & 0.78 \\
SuMo++ SuMo & 0.73 & 0.57 & 0.53 & 0.80 & 0.49 & 0.96 & 0.80 \\
SuMo++ BCE & 0.72 & \textbf{0.64} & \textbf{0.55} & 0.79 & \textbf{0.52} & 0.91 & 0.78 \\
\end{tabular}
\caption{
The mean of all over time integrated scores and the concordance index for each model and the relevant test dataset. The higher, the better. We only list our BCE loss and the SuMo loss if applicable. The CoxDeepNN model is also known as DeepHit or DeepSurv. For the individual scores see the appendix, Tables~\ref{tab:stats_A}--~\ref{tab:stats_iibs}.
}
\label{tab:stats_Mean}
\end{table*}

In Section~\ref{sec:separate_scores} of the appendix, we present tables with each separate integrated quality score, along with the non-integrated versions of the balanced accuracy and the $F_1$ score for each dataset in Section~\ref{sec:plots_non_integrated}. We chose these two scores as they are highly correlated with the mean of the scores and complement each other.

Table~\ref{tab:stats_Mean} shows that SuMo, SuMo+, and SuMo++ outperform all other models. For example, plots of survival curves see Section~\ref{sec:predicted_survival_curves} in the appendix. In particular, SuMo+ and SuMo++ slightly outperform SuMo in most cases while guaranteeing the output to be a survival curve. For the losses, the BCE loss tends to outperform the SuMo loss on survival datasets containing right-censoring; for more on the SuMo loss, see the appendix, Section~\ref{sec:training_details}, and~\cite{rindt2022survival}.

In contrast, the SuMo loss outperforms the BCE for the adapted regression datasets. We are not sure why that is, but the SuMo, unlike the BCE loss, is conditioned on whether the event was observed. This might decrease its performance if censoring is not independent of the time of the event. This would not be an issue for an entirely uncensored dataset.
\section{Conclusion}
In this paper, we accomplished four main objectives: 1. We examined how SA can bridge the well-studied domains of classification and regression. 2. We proposed a novel loss function. 3. We present the ﬁrst DL model that provably produces survival curves in continuous time and is also a universal approximator. 4. We suggested a more nuanced evaluation of survival models by modifying the integrated Brier score with classifier metrics.

We also retrofitted our new models to the well-established Cox model and its time-dependent version. Further we conducted a comprehensive comparison of recent and traditional methods. Our novel BCE loss outperforms the SuMo loss and the classical setup on classical survival tasks. In contrast, the SuMo loss outperforms on reinterpreted regression tasks where no censoring is present.

Our work presents the potential for more powerful survival models in fields such as clinical data, engineering, economics, and sociology. Therefore we hope our reinterpretation and stronger models -- applicable to modalities like images -- open up SA to a broader audience.

While new DL approaches offer promising results, the quality of uncertainty estimation remains an open area of research. For example, our BCE loss is influenced by $\sigma$ and distributions $\mathcal{T}_\pm$ in~\eqref{eq:T_minus} and \eqref{eq:T_plus}, which can affect the final model. However, this can also be an opportunity to adapt the model's predictions based on the specific application and evaluation criteria.
Finally, we want to point out that any survival model produces biased predictions if the censoring assumption, usually non-informative censoring, is violated.

\FloatBarrier
\acks{
There is no direct funding for this study, but the authors are grateful for the following indirect funding: The EU/EFPIA Innovative Medicines Initiative project DRAGON (101005122) (M.R., S.D., AIX-COVNET, C.-B.S.), the Trinity Challenge (M.R., C.-B.S.), the EPSRC Cambridge Mathematics of Information in Healthcare Hub EP/T017961/1 (M.R., S.D., J.H.F.R., J.A.D.A, C.-B.S.), the Cantab Capital Institute for the Mathematics of Information (C.-B.S.), the European Research Council under the European Union’s Horizon 2020 research and innovation programme grant agreement no. 777826 (C.-B.S.), the Alan Turing Institute (C.-B.S.), Wellcome Trust (J.H.F.R.), Cancer Research UK Cambridge Centre (C9685/A25177) (C.-B.S.), British Heart Foundation (J.H.F.R.), the NIHR Cambridge Biomedical Research Centre (J.H.F.R.), HEFCE (J.H.F.R.). In addition, C.-B.S. acknowledges support from the Leverhulme Trust project on ‘Breaking the non-convexity barrier’, the Philip Leverhulme Prize, the EPSRC grants EP/S026045/1 and EP/T003553/1 and the Wellcome Innovator Award RG98755. Finally, the AIX-COVNET collaboration is also grateful to Intel for financial support.\\

We also want to acknowledge and thank the members of the AIX-COVNET collaboration:
Michael Roberts$^{1}$, S{\"{o}}ren Dittmer$^{1,6}$, Ian Selby$^{7}$, Anna Breger$^{1,8}$, Matthew Thorpe$^{9}$, Julian Gilbey$^{1}$, Jonathan R. Weir-McCall$^{7,10}$, Effrossyni Gkrania-Klotsas$^{3}$, Anna Korhonen$^{11}$, Emily Jefferson$^{12}$, Georg Langs$^{13}$, Guang Yang$^{14}$, Helmut Prosch$^{13}$, Jacobus Preller$^{3}$, Jan Stanczuk$^{1}$, Jing Tang$^{15}$, Judith Babar$^{3}$, Lorena Escudero Sánchez$^{7}$, Philip Teare$^{16}$, Mishal Patel$^{16,17}$, Marcel Wassin$^{18}$, Markus Holzer$^{18}$, Nicholas Walton$^{19}$, Pietro Li{\'{o}}$^{20}$, Tolou Shadbahr$^{15}$, James H. F. Rudd$^{4}$, John A.D. Aston$^{5}$, Evis Sala$^{7}$ and Carola-Bibiane Schönlieb$^{1}$.\\

\noindent
${}^{1}$ Department of Applied Mathematics and Theoretical Physics, University of Cambridge, Cambridge, UK
${}^{2}$ A list of authors and their affiliations appears at the end of the paper
${}^{3}$ Addenbrooke’s Hospital, Cambridge University Hospitals NHS Trust, Cambridge, UK.
${}^{4}$ Department of Medicine, University of Cambridge, Cambridge, UK
${}^{5}$ Department of Pure Mathematics and Mathematical Statistics, University of Cambridge, Cambridge, UK
${}^{6}$ ZeTeM, University of Bremen, Bremen, Germany
${}^{7}$ Department of Radiology, University of Cambridge, Cambridge, UK
${}^{8}$ Faculty of Mathematics, University of Vienna, Austria.
${}^{9}$ Department of Mathematics, University of Manchester, Manchester, UK.
${}^{10}$ Royal Papworth Hospital, Cambridge, Royal Papworth Hospital NHS Foundation Trust, Cambridge, UK
${}^{11}$ Language Technology Laboratory, University of Cambridge, Cambridge, UK.
${}^{12}$ Population Health and Genomics, School of Medicine, University of Dundee, Dundee, UK.
${}^{13}$ Department of Biomedical Imaging and Image-guided Therapy, Computational Imaging Research Lab Medical University of Vienna, Vienna, Austria.
${}^{14}$ National Heart and Lung Institute, Imperial College London, London, UK.
${}^{15}$ Research Program in Systems Oncology, Faculty of Medicine, University of Helsinki, Helsinki, Finland.
${}^{16}$ Data Science \& Artificial Intelligence, AstraZeneca, Cambridge, UK.
${}^{17}$ Clinical Pharmacology \& Safety Sciences, AstraZeneca, Cambridge, UK.
${}^{18}$ contextflow GmbH, Vienna, Austria. 
${}^{19}$ Institute of Astronomy, University of Cambridge, Cambridge, UK. 
${}^{20}$ Department of Computer Science and Technology, University of Cambridge, Cambridge, UK.
}

\FloatBarrier
\bibliography{references}

\begin{thebibliography}{53}
\providecommand{\natexlab}[1]{#1}
\providecommand{\url}[1]{\texttt{#1}}
\expandafter\ifx\csname urlstyle\endcsname\relax
  \providecommand{\doi}[1]{doi: #1}\else
  \providecommand{\doi}{doi: \begingroup \urlstyle{rm}\Url}\fi

\bibitem[Akiba et~al.(2019)Akiba, Sano, Yanase, Ohta, and
  Koyama]{akiba2019optuna}
Takuya Akiba, Shotaro Sano, Toshihiko Yanase, Takeru Ohta, and Masanori Koyama.
\newblock {Optuna: A next-generation hyperparameter optimization framework}.
\newblock In \emph{{Proceedings of the 25th ACM SIGKDD international conference
  on knowledge discovery \& data mining}}, pages 2623--2631, 2019.

\bibitem[Bennett(1983)]{bennett1983log}
Steve Bennett.
\newblock {Log-logistic regression models for survival data}.
\newblock \emph{{Journal of the Royal Statistical Society: Series C (Applied
  Statistics)}}, 32\penalty0 (2):\penalty0 165--171, 1983.

\bibitem[Chilinski and Silva(2020)]{chilinski2020neural}
Pawel Chilinski and Ricardo Silva.
\newblock {Neural likelihoods via cumulative distribution functions}.
\newblock In \emph{{Conference on Uncertainty in Artificial Intelligence}},
  pages 420--429. PMLR, 2020.

\bibitem[Ching et~al.(2018)Ching, Zhu, and Garmire]{ching2018cox}
Travers Ching, Xun Zhu, and Lana~X Garmire.
\newblock {Cox-nnet: an artificial neural network method for prognosis
  prediction of high-throughput omics data}.
\newblock \emph{{PLoS computational biology}}, 14\penalty0 (4):\penalty0
  e1006076, 2018.

\bibitem[Cox(1972)]{cox1972regression}
David~R Cox.
\newblock {Regression models and life-tables}.
\newblock \emph{{Journal of the Royal Statistical Society: Series B
  (Methodological)}}, 34\penalty0 (2):\penalty0 187--202, 1972.

\bibitem[Cox and Oakes(2018)]{cox2018analysis}
David~Roxbee Cox and David Oakes.
\newblock \emph{{Analysis of survival data}}.
\newblock {Chapman and Hall/CRC}, 2018.

\bibitem[Crooks et~al.(2020)Crooks, West, Fogarty, Morling, Grainge, Gonem,
  Simmonds, Race, Juurlink, Briggs, et~al.]{crooks2020predicting}
Colin~J Crooks, Joe West, Andrew Fogarty, Joanne~R Morling, Matthew~J Grainge,
  Sherif Gonem, Mark Simmonds, Andrea Race, Irene Juurlink, Steve Briggs,
  et~al.
\newblock {Predicting the need for escalation of care or death from repeated
  daily clinical observations and laboratory results in patients with
  SARS-CoV-2 during 2020: a retrospective population-based cohort study from
  the United Kingdom}.
\newblock \emph{{medRxiv}}, 2020.

\bibitem[Crowley and Breslow(1984)]{crowley1984statistical}
John Crowley and Norman Breslow.
\newblock {Statistical analysis of survival data}.
\newblock \emph{{Annual review of public health}}, 5\penalty0 (1):\penalty0
  385--411, 1984.

\bibitem[Danks and Yau(2022)]{danks2022derivative}
Dominic Danks and Christopher Yau.
\newblock {Derivative-Based Neural Modelling of Cumulative Distribution
  Functions for Survival Analysis}.
\newblock In \emph{{International Conference on Artificial Intelligence and
  Statistics}}, pages 7240--7256. {PMLR}, 2022.

\bibitem[Davidson-Pilon(2022)]{davidson_pilon_cameron_2022_7111973}
Cameron Davidson-Pilon.
\newblock {lifelines, survival analysis in Python}, September 2022.
\newblock URL \url{https://doi.org/10.5281/zenodo.7111973}.
\newblock {If you use this software, please cite it using these metadata.}

\bibitem[Du(2021)]{du2021beyond}
Hailiang Du.
\newblock {Beyond Strictly Proper Scoring Rules: The Importance of Being
  Local}.
\newblock \emph{{Weather and Forecasting}}, 36\penalty0 (2):\penalty0 457--468,
  2021.

\bibitem[Efron(1988)]{efron1988logistic}
Bradley Efron.
\newblock {Logistic regression, survival analysis, and the Kaplan-Meier curve}.
\newblock \emph{{Journal of the American statistical Association}}, 83\penalty0
  (402):\penalty0 414--425, 1988.

\bibitem[Fisher et~al.(1999)Fisher, Lin, et~al.]{fisher1999time}
Lloyd~D Fisher, Danyu~Y Lin, et~al.
\newblock {Time-dependent covariates in the Cox proportional-hazards regression
  model}.
\newblock \emph{{Annual review of public health}}, 20\penalty0 (1):\penalty0
  145--157, 1999.

\bibitem[Gamel and McLean(1994)]{gamel1994stable}
John~W Gamel and Ian~W McLean.
\newblock {A stable, multivariate extension of the log-normal survival model}.
\newblock \emph{{Computers and Biomedical Research}}, 27\penalty0 (2):\penalty0
  148--155, 1994.

\bibitem[Giunchiglia et~al.(2018)Giunchiglia, Nemchenko, and
  Schaar]{giunchiglia2018rnn}
Eleonora Giunchiglia, Anton Nemchenko, and Mihaela van~der Schaar.
\newblock {RNN-SURV: A deep recurrent model for survival analysis}.
\newblock In \emph{{International conference on artificial neural networks}},
  pages 23--32. {Springer}, 2018.

\bibitem[Graf et~al.(1999)Graf, Schmoor, Sauerbrei, and
  Schumacher]{graf1999assessment}
Erika Graf, Claudia Schmoor, Willi Sauerbrei, and Martin Schumacher.
\newblock {Assessment and comparison of prognostic classification schemes for
  survival data}.
\newblock \emph{{Statistics in medicine}}, 18\penalty0 (17-18):\penalty0
  2529--2545, 1999.

\bibitem[Han et~al.(2016)Han, Lee, Lee, Kim, Son, Yang, Lee, and
  Kim]{han2016oasis}
Seong~Kyu Han, Dongyeop Lee, Heetak Lee, Donghyo Kim, Heehwa~G Son, Jae-Seong
  Yang, Seung-Jae~V Lee, and Sanguk Kim.
\newblock Oasis 2: online application for survival analysis 2 with features for
  the analysis of maximal lifespan and healthspan in aging research.
\newblock \emph{Oncotarget}, 7\penalty0 (35):\penalty0 56147, 2016.

\bibitem[He et~al.(2015)He, Zhang, Ren, and Sun]{he2015delving}
Kaiming He, Xiangyu Zhang, Shaoqing Ren, and Jian Sun.
\newblock {Delving deep into rectifiers: Surpassing human-level performance on
  imagenet classification}.
\newblock In \emph{{Proceedings of the IEEE international conference on
  computer vision}}, pages 1026--1034, 2015.

\bibitem[He et~al.(2016)He, Zhang, Ren, and Sun]{he2016deep}
Kaiming He, Xiangyu Zhang, Shaoqing Ren, and Jian Sun.
\newblock Deep residual learning for image recognition.
\newblock In \emph{Proceedings of the IEEE conference on computer vision and
  pattern recognition}, pages 770--778, 2016.

\bibitem[Ioffe and Szegedy(2015)]{ioffe2015batch}
Sergey Ioffe and Christian Szegedy.
\newblock {Batch normalization: Accelerating deep network training by reducing
  internal covariate shift}.
\newblock In \emph{{International conference on machine learning}}, pages
  448--456. {PMLR}, 2015.

\bibitem[Kalbfleisch and Prentice(2011)]{kalbfleisch2011statistical}
J.D. Kalbfleisch and R.L. Prentice.
\newblock \emph{{The Statistical Analysis of Failure Time Data}}.
\newblock {Wiley Series in Probability and Statistics}. {Wiley}, 2011.
\newblock ISBN 9781118031230.
\newblock URL \url{https://books.google.de/books?id=BR4Kq-a1MIMC}.

\bibitem[Kaplan and Meier(1958)]{kaplan1958nonparametric}
Edward~L Kaplan and Paul Meier.
\newblock {Nonparametric estimation from incomplete observations}.
\newblock \emph{{Journal of the American statistical association}}, 53\penalty0
  (282):\penalty0 457--481, 1958.

\bibitem[Katzman et~al.(2018)Katzman, Shaham, Cloninger, Bates, Jiang, and
  Kluger]{katzman2018deepsurv}
Jared~L Katzman, Uri Shaham, Alexander Cloninger, Jonathan Bates, Tingting
  Jiang, and Yuval Kluger.
\newblock {DeepSurv: personalized treatment recommender system using a Cox
  proportional hazards deep neural network}.
\newblock \emph{{BMC medical research methodology}}, 18\penalty0 (1):\penalty0
  1--12, 2018.

\bibitem[Kingma and Ba(2014)]{kingma2014adam}
Diederik~P Kingma and Jimmy Ba.
\newblock {Adam: A method for stochastic optimization}.
\newblock \emph{arXiv preprint arXiv:1412.6980}, 2014.

\bibitem[Kleinbaum et~al.(2012)Kleinbaum, Klein, et~al.]{kleinbaum2012survival}
David~G Kleinbaum, Mitchel Klein, et~al.
\newblock \emph{{Survival analysis: a self-learning text}}, volume~3.
\newblock {Springer}, 2012.

\bibitem[Kong et~al.(2018)Kong, Ibrahim, Lee, and Zhu]{kong2018flcrm}
Dehan Kong, Joseph~G Ibrahim, Eunjee Lee, and Hongtu Zhu.
\newblock Flcrm: Functional linear cox regression model.
\newblock \emph{Biometrics}, 74\penalty0 (1):\penalty0 109--117, 2018.

\bibitem[Kvamme et~al.(2019)Kvamme, Borgan, and Scheel]{kvamme2019time}
H{\aa}vard Kvamme, {\O}rnulf Borgan, and Ida Scheel.
\newblock {Time-to-event prediction with neural networks and Cox regression}.
\newblock \emph{arXiv preprint arXiv:1907.00825}, 2019.

\bibitem[Lai et~al.(2006)Lai, Murthy, and Xie]{lai2006weibull}
Chin-Diew Lai, DN~Murthy, and Min Xie.
\newblock {Weibull distributions and their applications}.
\newblock In \emph{{Springer Handbooks}}, pages 63--78. Springer, 2006.

\bibitem[Lang(2005)]{lang2005monotonic}
Bernhard Lang.
\newblock {Monotonic multi-layer perceptron networks as universal
  approximators}.
\newblock In \emph{{International conference on artificial neural networks}},
  pages 31--37. {Springer}, 2005.

\bibitem[Lee et~al.(2018)Lee, Zame, Yoon, and Van Der~Schaar]{lee2018deephit}
Changhee Lee, William Zame, Jinsung Yoon, and Mihaela Van Der~Schaar.
\newblock {Deephit: A deep learning approach to survival analysis with
  competing risks}.
\newblock In \emph{{Proceedings of the AAAI conference on artificial
  intelligence}}, volume~32, 2018.

\bibitem[Lee et~al.(2015)Lee, Zhu, Kong, Wang, Giovanello, and
  Ibrahim]{lee2015bflcrm}
Eunjee Lee, Hongtu Zhu, Dehan Kong, Yalin Wang, Kelly~Sullivan Giovanello, and
  Joseph~G Ibrahim.
\newblock Bflcrm: A bayesian functional linear cox regression model for
  predicting time to conversion to alzheimer’s disease.
\newblock \emph{The annals of applied statistics}, 9\penalty0 (4):\penalty0
  2153, 2015.

\bibitem[Lemeshow et~al.(2011)Lemeshow, May, and
  Hosmer~Jr]{lemeshow2011applied}
Stanley Lemeshow, Susanne May, and David~W Hosmer~Jr.
\newblock \emph{{Applied survival analysis: regression modeling of
  time-to-event data}}.
\newblock {John Wiley \& Sons}, 2011.

\bibitem[Nicolau et~al.(2011)Nicolau, Levine, and
  Carlsson]{nicolau2011topology}
Monica Nicolau, Arnold~J Levine, and Gunnar Carlsson.
\newblock {Topology based data analysis identifies a subgroup of breast cancers
  with a unique mutational profile and excellent survival}.
\newblock \emph{{Proceedings of the National Academy of Sciences}},
  108\penalty0 (17):\penalty0 7265--7270, 2011.

\bibitem[Pace and Barry(1997)]{pace1997sparse}
R~Kelley Pace and Ronald Barry.
\newblock {Sparse spatial autoregressions}.
\newblock \emph{{Statistics \& Probability Letters}}, 33\penalty0 (3):\penalty0
  291--297, 1997.

\bibitem[Paszke et~al.(2017)Paszke, Gross, Chintala, Chanan, Yang, DeVito, Lin,
  Desmaison, Antiga, and Lerer]{paszke2017automatic}
Adam Paszke, Sam Gross, Soumith Chintala, Gregory Chanan, Edward Yang, Zachary
  DeVito, Zeming Lin, Alban Desmaison, Luca Antiga, and Adam Lerer.
\newblock {Automatic differentiation in pytorch}.
\newblock \emph{-}, 2017.

\bibitem[Ren et~al.(2019)Ren, Qin, Zheng, Yang, Zhang, Qiu, and
  Yu]{ren2019deep}
Kan Ren, Jiarui Qin, Lei Zheng, Zhengyu Yang, Weinan Zhang, Lin Qiu, and Yong
  Yu.
\newblock {Deep recurrent survival analysis}.
\newblock In \emph{{Proceedings of the AAAI Conference on Artificial
  Intelligence}}, volume~33, pages 4798--4805, 2019.

\bibitem[Rindt et~al.(2022)Rindt, Hu, Steinsaltz, and
  Sejdinovic]{rindt2022survival}
David Rindt, Robert Hu, David Steinsaltz, and Dino Sejdinovic.
\newblock {Survival regression with proper scoring rules and monotonic neural
  networks}.
\newblock In \emph{{International Conference on Artificial Intelligence and
  Statistics}}, pages 1190--1205. PMLR, 2022.

\bibitem[Royston and Parmar(2002)]{royston2002flexible}
Patrick Royston and Mahesh~KB Parmar.
\newblock {Flexible parametric proportional-hazards and proportional-odds
  models for censored survival data, with application to prognostic modelling
  and estimation of treatment effects}.
\newblock \emph{{Statistics in medicine}}, 21\penalty0 (15):\penalty0
  2175--2197, 2002.

\bibitem[Santurkar et~al.(2018)Santurkar, Tsipras, Ilyas, and
  Madry]{santurkar2018does}
Shibani Santurkar, Dimitris Tsipras, Andrew Ilyas, and Aleksander Madry.
\newblock {How does batch normalization help optimization?}
\newblock \emph{{Advances in neural information processing systems}}, 31, 2018.

\bibitem[Sauerbrei and Royston(1999)]{sauerbrei1999building}
Willi Sauerbrei and Patrick Royston.
\newblock {Building multivariable prognostic and diagnostic models:
  transformation of the predictors by using fractional polynomials}.
\newblock \emph{{Journal of the Royal Statistical Society: Series A (Statistics
  in Society)}}, 162\penalty0 (1):\penalty0 71--94, 1999.

\bibitem[Saul(2016)]{saul2016gaussian}
Alan~D Saul.
\newblock \emph{{Gaussian process based approaches for survival analysis}}.
\newblock PhD thesis, {University of Sheffield}, 2016.

\bibitem[Schmoor et~al.(2000)Schmoor, Sauerbrei, Bastert, Schumacher, and
  Group]{schmoor2000role}
Claudia Schmoor, Willi Sauerbrei, Gunter Bastert, Martin Schumacher, and German
  Breast Cancer~Study Group.
\newblock {Role of isolated locoregional recurrence of breast cancer: results
  of four prospective studies}.
\newblock \emph{{Journal of Clinical Oncology}}, 18\penalty0 (8):\penalty0
  1696--1708, 2000.

\bibitem[Schwab et~al.(2021)Schwab, Mehrjou, Parbhoo, Celi, Hetzel, Hofer,
  Sch{\"o}lkopf, and Bauer]{schwab2021real}
Patrick Schwab, Arash Mehrjou, Sonali Parbhoo, Leo~Anthony Celi, J{\"u}rgen
  Hetzel, Markus Hofer, Bernhard Sch{\"o}lkopf, and Stefan Bauer.
\newblock {Real-time prediction of COVID-19 related mortality using electronic
  health records}.
\newblock \emph{{Nature communications}}, 12\penalty0 (1):\penalty0 1--16,
  2021.

\bibitem[Shahin et~al.(2022)Shahin, Jacob, Alexander, and
  Barber]{shahin2022survival}
Ahmed~H Shahin, Joseph Jacob, Daniel~C Alexander, and David Barber.
\newblock {Survival Analysis for Idiopathic Pulmonary Fibrosis using CT Images
  and Incomplete Clinical Data}.
\newblock \emph{arXiv preprint arXiv:2203.11391}, 2022.

\bibitem[Sonabend(2021)]{sonabend2021theoretical}
Raphael Edward~Benjamin Sonabend.
\newblock \emph{{A theoretical and methodological framework for machine
  learning in survival analysis: Enabling transparent and accessible predictive
  modelling on right-censored time-to-event data}}.
\newblock PhD thesis, {UCL (University College London)}, 2021.

\bibitem[Verma(2019)]{clocks}
Shiva Verma.
\newblock {Analog-Clocks dataset}.
\newblock \url{https://www.kaggle.com/datasets/shivajbd/analog-clocks}, 2019.

\bibitem[Wang et~al.(2019)Wang, Li, and Reddy]{wang2019machine}
Ping Wang, Yan Li, and Chandan~K Reddy.
\newblock Machine learning for survival analysis: A survey.
\newblock \emph{ACM Computing Surveys (CSUR)}, 51\penalty0 (6):\penalty0 1--36,
  2019.

\bibitem[Wiegand et~al.(2022)Wiegand, Cowan, Waddington, Halsall, Keevil, Tom,
  Taylor, Gkrania-Klotsas, Preller, and Goudie]{wiegand2022development}
Martin Wiegand, Sarah~L Cowan, Claire~S Waddington, David~J Halsall, Victoria~L
  Keevil, Brian~DM Tom, Vince Taylor, Effrossyni Gkrania-Klotsas, Jacobus
  Preller, and Robert~JB Goudie.
\newblock {Development and validation of a dynamic 48-hour in-hospital
  mortality risk stratification for COVID-19 in a UK teaching hospital: a
  retrospective cohort study}.
\newblock \emph{{medRxiv}}, pages 2021--02, 2022.

\bibitem[Yang et~al.(2018)Yang, Cai, and Reddy]{yang2018spatio}
Guolei Yang, Ying Cai, and Chandan~K Reddy.
\newblock Spatio-temporal check-in time prediction with recurrent neural
  network based survival analysis.
\newblock In \emph{Proceedings of the Twenty-Seventh International Joint
  Conference on Artificial Intelligence}, 2018.

\bibitem[Yang et~al.(2011)Yang, Nam, Seo, Han, Choi, Nam, Lee, and
  Kim]{yang2011oasis}
Jae-Seong Yang, Hyun-Jun Nam, Mihwa Seo, Seong~Kyu Han, Yonghwan Choi, Hong~Gil
  Nam, Seung-Jae Lee, and Sanguk Kim.
\newblock Oasis: online application for the survival analysis of lifespan
  assays performed in aging research.
\newblock \emph{PloS one}, 6\penalty0 (8):\penalty0 e23525, 2011.

\bibitem[Youden(1950)]{youden1950index}
William~J Youden.
\newblock {Index for rating diagnostic tests}.
\newblock \emph{{Cancer}}, 3\penalty0 (1):\penalty0 32--35, 1950.

\bibitem[Zhang et~al.(2022)Zhang, Wong, Mann, Muller, and
  Yang]{zhang2022survbenchmark}
Yunwei Zhang, Germaine Wong, Graham Mann, Samuel Muller, and Jean~YH Yang.
\newblock {SurvBenchmark: comprehensive benchmarking study of survival analysis
  methods using both omics data and clinical data}.
\newblock \emph{{GigaScience}}, 11, 2022.

\bibitem[Zhong and Tibshirani(2019)]{zhong2019survival}
Chenyang Zhong and Robert Tibshirani.
\newblock {Survival analysis as a classification problem}.
\newblock \emph{arXiv preprint arXiv:1909.11171}, 2019.

\end{thebibliography}

\FloatBarrier
\newpage
\appendix
\section{The MONDE and SuMo networks}
\label{sec:monde_sumo}
We begin with the MONDE network~\cite{chilinski2020neural}, which allows one to parameterize a mapping
\begin{equation}
    M:\mathbb{R}^m \ni z \mapsto M(z) \in \mathbb{R}
\end{equation}
that is guaranteed to be monotonically increasing. It is a feed-forward neural network with each layer being of the form
\begin{equation}
    z_{k+1} = \varphi_k\left(W_kz_k\right).
\end{equation}
Here $z_0=z$ and $W_k$ is an affine map with the weights of the linear part constrained to be non-negative. For the last layer, outputting a scalar, they set $\varphi_K=\mbox{id}$, otherwise $\varphi_k=\tanh$.

The SuMo network~\cite{rindt2022survival} consists of MONDE and a feature-extracting network
\begin{equation}
    Q: \mathbb{R}^n \ni x \mapsto Q(x) \in \mathbb{R}^{k}
\end{equation}
combined with
\begin{equation}
    \mathscr{S}: \mathbb{R} \times \mathbb{R}^n \mapsto [0,1]
\end{equation}
to give
\begin{equation}
    \mathscr{S}(t, x) = 1 - \mbox{sigmoid}\left(M([t, Q(x)])\right).
\end{equation}
Here $[t, Q(x)]$ denotes the concatenation of $t$ and $Q(x).$ As $M$ and the $\mbox{sigmoid}$ function are monotonically increasing, the SuMo network, $\mathscr{S}(t, x)$, is monotonically decreasing in $t$.

As discussed in~\cite{rindt2022survival, chilinski2020neural}, the SuMo network is not only monotonically decreasing but is also a universal approximator of survival curves. 
However, as there is no guarantee that $\mathscr{S}(0, x)=1$ for all $x \in \mathbb{R}^n$, it does not necessarily produce survival curves. Note, for the SuMo model, $\mathscr{S}(0, x)=1$ is even impossible, as $\mbox{sigmoid}(\cdot) > 0.$

\section{Initialization of MONDE and MONDE+}
\label{sec:initalization}
We used Optuna~\cite{akiba2019optuna} to run $1,000$ hyperparameter optimization steps on the task in Section~\ref{sec:toy_example} to optimize the scale and sign of the random initializations of MONDE and MONDE+. This was motivated by the initialization's strong influence demonstrated by the histograms in Figure~\ref{fig:hists}.

\subsection{Default MONDE initialization}
Following~\cite{rindt2022survival}, the default MONDE initialization initializes both the weight matrix and the bias with Gaussian noise with mean zero. The standard deviation used in the initial Gaussian noise for the  weight matrix is $\sqrt{\text{input size}}$ and for the biases $\sqrt{\text{output size}}$ is used.

\subsection{Hyperparameter optimization for MONDE}
Here we take the absolute value of the default weights described above and scale them by a factor from the interval $[-10,10]$. Performing 1000 hyperparameter optimization steps yields the factor $4.6$ for the matrix and $6.6$ for the bias.

The hyperparameter objective function was given by the maximum loss of three training runs, as described in Section~\ref{sec:toy_example}. We took the maximum loss of three runs in the hope of improving the robustness to ``unlucky'' initializations, as shown by the histograms in Figure~\ref{fig:hists}.

\subsection{Hyperparameter optimization for MONDE+}
For the convenience of the reader, we recall the definition of a MONDE+ layer to be
\begin{equation}
    z_{k+1}(t, z_k, z_0) = H_kz_k + \sigma_k(\tilde z_k(t, z_k) + B_kz_k + L_kz_0)
\end{equation}
with
\begin{equation}
    \tilde z_k(t, z_k) = A_k\left(\phi_k(a_kt + b_k) \circ \psi_k(G_kz_k)\right)
\end{equation}
where the capital letters denote affine maps and $\circ$ denotes the Hadamard product.

For simplicity we initialized $a_k\in\mathbb{R}^{64}$ with the uniform distribution between $0$ and $1$ and $b_k\in\mathbb{R}^{64}$ to be zero.

The initialization of all matrices and biases in the affine maps was optimized before training the models. We first initialized with the Kaiming uniform initialization~\cite{he2015delving} as implemented by PyTorch~\cite{paszke2017automatic}. With the exception of $L_k$, we took their absolute values and scaled them by a factor (the hyperparameter) between $[0, 10]$, as these matrices had to be non-negative.

As in the MONDE case, we also scaled the biases as purely non-negative or non-positive. We did so to mediate the following effect. Some of the matrices need to be non-negative, which, as we found, can lead to impaired training performance. This is likely due to a substantial distribution shift in the activations over the course of multiple layers making the optimization landscape less smooth~\cite{ioffe2015batch, santurkar2018does}. It seems that having accordingly biased biases can counter this effect to some degree.

We again optimized the same hyperparameter objective as in the MONDE case above for 1000 optimization steps. For simplicity, we optimized the factors of the matrices of all affine mappings as one, which yielded the scaling factor $0.2$. The resulting factors for the biases were $-8.5$ for $B_k$, $10$ for $G_k$, the bias of $H_k$ given by the Kaiming initialization was used, and for $A_k$ we removed the bias entirely, as it would be redundant.

\section{Proof of Lemma~\ref{lemma:monde_plus}}
\label{sec:monde_proof}
As the concatenation of monotonically increasing functions is monotonically increasing, it suffices to prove the monotonicity of a single layer.

We will prove monotonicity in $t$ by showing that the derivative of the output of a layer
\begin{equation}
    z_{k+1}(t, z_k, z_0) = H_kz_k + \sigma_k(\tilde z_k(t, z_k) + B_kz_k + L_kz_0)
\end{equation}
is positive in\ $t$ and $z_k$.

First, we have that
\begin{align}
    \begin{split}
        \partial_t \tilde z_k(t, z_k)
        &=  \overline A_k\partial_t \left[\phi_k(a_kt + b_k) \circ \psi_k(G_kz_k)\right]\\
        &=  \overline A_k \{\partial_t[\phi_k(a_kt + b_k)] \circ \psi_k(G_kz_k) \\& + \phi_k(a_kt + b_k) \circ \partial_t[\psi_k(G_kz_k)]\}\\
        &=  \overline A_k \{[\nabla\phi_k](a_kt + b_k) \circ a_k \circ \psi_k(G_kz_k) \\& + \phi_k(a_kt + b_k) \circ \partial_t[\psi_k(G_kz_k)]\}\\
        &=  \overline A_k \{[\nabla\phi_k](a_kt + b_k) \circ a_k \circ \psi_k(G_kz_k) \\& + \phi_k(a_kt + b_k) \circ [\partial_t\psi_k](G_kz_k) \circ \overline G_k\partial_tz_k\}\\
    \end{split}
\end{align}
and
\begin{align}
    \begin{split}
        \partial_{z_k} \tilde z_k(t, z_k)
        &= \overline A_k\partial_{z_k} \left[\phi_k(a_kt + b_k) \circ \psi_k(G_kz_k)\right]\\
        &= \overline A_k \left[\phi_k(a_kt + b_k) \circ \partial_{z_k}[\psi_k(G_kz_k)\right]\\
        &= \overline A_k \left[D\{\phi_k(a_kt + b_k)\} D\{(\nabla \psi_k)(G_kz_k)\} \overline G_k\right]\\
    \end{split}
\end{align}
where $D\{c\}$ denotes the diagonal matrix associated with a vector $c$. Also, $\overline A_k$ and $\overline G_k$ are the linear parts of the affine operators $A_k$ and $G_k$ respectively. As they only have non-negative weights in their linear parts and the non-linearities, $\phi_k$, and $\psi_k$, are non-negative and monotonically increasing, both derivatives are pointwise non-negative.

Using this, we can compute the corresponding derivates of $z_k.$
We have
\begin{align}
    \begin{split}
        &\partial_t z_{k+1}(t, z_k, z_0)
        = H_k\partial_tz_k + (\nabla\sigma_k)(\tilde z_k(t, z_k) \\&+ B_kz_k + L_kz_0) \circ (\partial_t\tilde z_k(t, z_k) + B_k\partial_tz_k)
    \end{split}
\end{align}
and
\begin{align}
    \begin{split}
        &\partial_{z_k} z_{k+1}(t, z_k, z_0)
        = H_k + D\{(\nabla\sigma_k)(\tilde z_k(t, z_k)\\ &+ B_kz_k + L_kz_0)\} (\partial_{z_k}\tilde z_k(t, z_k) + B_k).
    \end{split}
\end{align}
For the same reasons as above, every term in these expressions is pointwise non-negative. $\square$

\section{Proof of Lemma~\ref{lemma:cOx}}
\label{sec:proof_cOx}
We begin by proving the first two points.
As $\alpha(t,x),\Lambda_0(t) \ge 0$ for all $(t,x)\in\mathbb{R}_{\ge0}\times\mathbb{R}^n$, it follows that $\mathscr{S}_\cOx(t, x)\in[0,1]$. Since $\Lambda_0(0)=0$, we have the maxima $\mathscr{S}_\cOx(0, x)=1$.

We will now show that $\mathscr{S}_\cOx$ is monotonically decreasing in $t$.
As the $\beta_i^\pm$ are monotonically increasing, the $\omega_i$ are monotonically increasing in $t$. Therefore $\alpha$ is monotonically increasing in $t$, which means $\alpha\Lambda_0$ is monotonically increasing in $t$ and therefore $\mathscr{S}_\cOx$ is monotonically decreasing in $t$.

The continuity in $t$ of $\omega_i$ follows from the continuity of the functions involved. The minima at $\omega_i(0, x)=\beta_i^+(0)$ follows from the fact, that $\beta_i^-(0)$ and $\beta_i^+(0)$ are both monotonically increasing and $\beta^-(0)=\beta^+(0)$.

We will now prove the continuity of $\alpha$ in $(t, x)$. We can reduce this problem to the continuity of each separate
\begin{equation}
    \omega_i(t, x)|x_i-o_i|.
\end{equation}
As all functions involved are otherwise continuous, we only need to show continuity at $x_i=o_i.$ Considering the limits of each of the two factors completes the proof. $\square$

\section{Datasets and preprocessing}
\label{sec:datasets_appendix}
For the NKI dataset, we reduced the number of features to make it accessible to classical survival models and easier to handle in general. We did so by only retaining only 9 of the features with a high correlation with survival time and observed outcome. This left us with the features: NM\_004701, NM\_001168, NM\_003430, grade, NM\_003981, AL050227, NM\_018410, NM\_000779, and NM\_001809.

We assembled the COVID-19 dataset from COVID-19 patient data from Addenbrooke's hospital in Cambridge, UK. We used the data of all patients on the day of their first positive COVID-19 test. We then used the following features of that day: age, gender, DNA CPR, median ventilation score over 24 hours, SpO2, pulse, temperature, respiration, NEWS2 score, O2 flow rate, FiO2, weight, height, BMI, Glasgow coma scale score, and SpO2/FiO2~\cite{wiegand2022development}. We used these variables either due to their ready availability or because they were shown to be useful in previous work~\cite{wiegand2022development}. We gathered their survival time via their death records. We assumed we had no censoring within the first 28 days and then marked all other patients as censored on the 28th day. Exact censoring was hard to perform, but the hospital transferred only a small number of ICU patients, therefore, we believe that for the analysis in this paper, censoring is a negligible problem in this datasets.

To avoid numerical problems for the classical models, we also divided the times of the GBSG2 and Lymph datasets by $100$ and those of the Recur dataset by $10$. Except for standardizing the mean and standard deviation to 0 and 1, respectively, we did not use any other preprocessing.

In addition to the original sources, one can find summaries of the Lymph, Recur, and GBSG2 datasets in the \textit{lifelines} documentation~\cite{davidson_pilon_cameron_2022_7111973}.

\section{Training details}
\label{sec:training_details}
\subsection{Training}

We used \textit{lifelines}~\cite{davidson_pilon_cameron_2022_7111973} to train the classical models, i.e. Kaplan-Meier \cite{kaplan1958nonparametric}, Log-Logistic \cite{bennett1983log}, Log-Normal \cite{gamel1994stable}, Weibull \cite{lai2006weibull}, and two Cox models \cite{cox2018analysis} (one based on a piecewise constant \cite{crowley1984statistical} and one on a spline \cite{efron1988logistic, royston2002flexible} baseline.)

For all the neural network models, we trained these using \textit{PyTorch}~\cite{paszke2017automatic}. As the network training is stochastic, we train each network setup five times and pick the best model based on the mean score over all evaluation metrics on the validation set.
For the neural networks, we also compare our BCE loss from~\eqref{eq:bce_loss} with the survival loss from~\cite{rindt2022survival}, namely
\begin{equation}
    L_\text{SuMo} = -\left[e \log f(t|x) + (1 - e) \log S(t|x)\right].
\end{equation}
For the convenience of the reader we recall $f = -\partial_t S(t|x)$. Following common practice, we will use a weighted version of the BCE in the BCE loss. To ensure fair comparability, we therefore introduce a hyperparameter $\gamma>0$ in the SuMo loss defining
\begin{equation}
    L_\text{SuMo}^\gamma = -\left[e \gamma\log f(t|x) + (1 - e) \log S(t|x)\right].
\end{equation}

For every dataset, except Clocks, we trained the networks CoxNN, \cOx NN, CoxDeepNN, SuMo, SuMo+, and SuMo++. For the Clocks dataset, we train SuMo, SuMo+, and SuMo++ as well as the CoxDeepNNs with the convolutional feature network and the Kaplan-Meier model. The CoxNN is based on Section~\ref{sec:cox_simple} and the \cOx NN is based on Section~\ref{sec:cox_time_dependent}. The CoxDeepNN is also based on Section~\ref{sec:cox_simple}, but the linear scalar product in~\eqref{eq:scalar_product_cox} is replaced by a deep feature network as independently proposed by~\cite{katzman2018deepsurv, kvamme2019time} and~\cite{shahin2022survival}. For the exact architecture of the feature networks of the CoxDeepNN and SuMo, SuMo+, and SuMo++, see the following section.

During our numerical experiments, we found that the training of any Cox-like model via $L_\text{SuMo}$ tends to diverge, especially if the training data is predominately non-censored. In those cases, the training with $L_\text{BCE}$ tends to be more stable but occasionally also diverges. We could ameliorate these issues by employing weight decay for all Cox-like models during all trainings.

All hyperparameters and training details for the different datasets are detailed in appendix Section~\ref{sec:app_training}.

\subsection{SuMo's, SuMo+'s, and SuMo++'s dense feature network}
All feature networks are simple feed-forward networks. SuMo, SuMo+, and SuMo++ have three dense layers of width 32, each followed by a ReLU activation and a layer-normalization.

\subsection{SuMo's, SuMo+'s, and SuMo++'s convolutional feature network}
\label{sec:sumo_feature_d}
In the convolutional setting, i.e., for the Clocks dataset, SuMo, SuMo+, and SuMo++ have three 2-dimensional convolution layers, each with 128 channels, a kernel size of 3, and a padding of 1. Each is followed by a max-pooling layer with a kernel size of 3, a stride of 2, and padding of 1. Each of these pooling layers is followed by a ReLU activation and a layer normalization. These three layer-blocks are followed by a dense layer of width 48, a ReLU, a layer-normalization, another dense layer of width 48, and a final ReLU.

\subsection{CoxDeepNN's dense feature network}
The network begins with two dense layers, the first with a width of twice the number of input features, see Table~\ref{tab:dataset_overview} and the second with a width of 8. Each of the two layers is followed by a layer normalization and a ReLU. Finally, the last layer is a dense layer of width 1.
 
\subsection{CoxDeepNN's convolutional feature network}
This network is the same as the one in Section~\ref{sec:sumo_feature_d}, except that the $\exp$ function replaces the final ReLU layer.

\subsection{Hyperparameters}
\label{sec:app_training}
For all networks, we use the Adam optimizer~\cite{kingma2014adam} with a step-size of $10^{-3}$, gradient clipping of $1$, and batch size of $8$. After each batch, we also evaluate the loss of a batch from the validation set. We stopped the training when the $512$-moving average over these validation batches did not decrease for $8192$ batches or after at most $200000$ training steps. 
With the exception of the Clocks dataset, which we trained on an Nvidia GeForce GTX 1080 Ti, we trained all models on the CPU. Each training took approximately $5$ to $90$ minutes.

For each dataset, we optimized the following hyperparameters with Optuna~\cite{akiba2019optuna} by evaluating the results of $100$ short training runs, each only $1024$ training steps long:
\begin{itemize}
    \item $\gamma$ for the $L_\text{SuMo}^\gamma$ loss over $[0,10]$.
    \item The BCE weight $L_\text{BCE}$ loss over $[0,1]$.
    \item The $\sigma$ for the $L_\text{BCE}$ loss indirectly via the factor $\sigma_\text{factor}\in[0,1]$ and $\sigma = \sigma_\text{factor} T_\text{max}$.
    \item One joint weight decay parameter for all Cox-like models in the case of the $L_\text{SuMo}^\gamma$ loss over $[0,1]$ and a separate one for the $L_\text{BCE}$ loss.
\end{itemize}
For the resulting parameters, see Figure~\ref{fig:hyperparamters}. We employed the mean of the concordance index and all integrated scores, except the integrated Brier score, as the loss function. The integrated Brier score was incorporated at a later stage of the paper. We determined appropriate ranges for the hyperparameters through a series of preliminary experiments.
\begin{table*}
    \centering
    \begin{tabular}{|c||c|c|c|c|c|}
        \hline
        Dataset & $\gamma$ & BCE weight & $\sigma_\text{factor}$ & weight decay for SuMo & weight decay for BCE\\
        \hline
        GBSG2 & 2.70 & 0.71 & 0.82 & 0.005 & 0.020\\
        Recur & 0.87 & 0.85 & 0.96 & 0.001 & 0.001\\
        NKI & 5.47 & 0.97 & 0.98 & 0.000 & 0.000\\
        Lymph & 3.44 & 0.86 & 0.79 & 0.004 & 0.002\\
        COVID & 2.49 & 0.92 & 0.71 & 0.000 & 0.002\\
        Clocks & 9.39 & 0.91 & 0.26 & 0.000 & 0.000\\
        California & 0.89 & 0.53 & 0.50 & 0.009 & 0.005\\
        \hline
    \end{tabular}
    \caption{Results of the hyperparameter search.}
    \label{fig:hyperparamters}
\end{table*}

\section{Further numerical results}
\label{sec:further_num_res}
\begin{figure}
    \centering
    \includegraphics[width=.56\textwidth, trim={30mm 60mm 35mm 55mm}, clip]{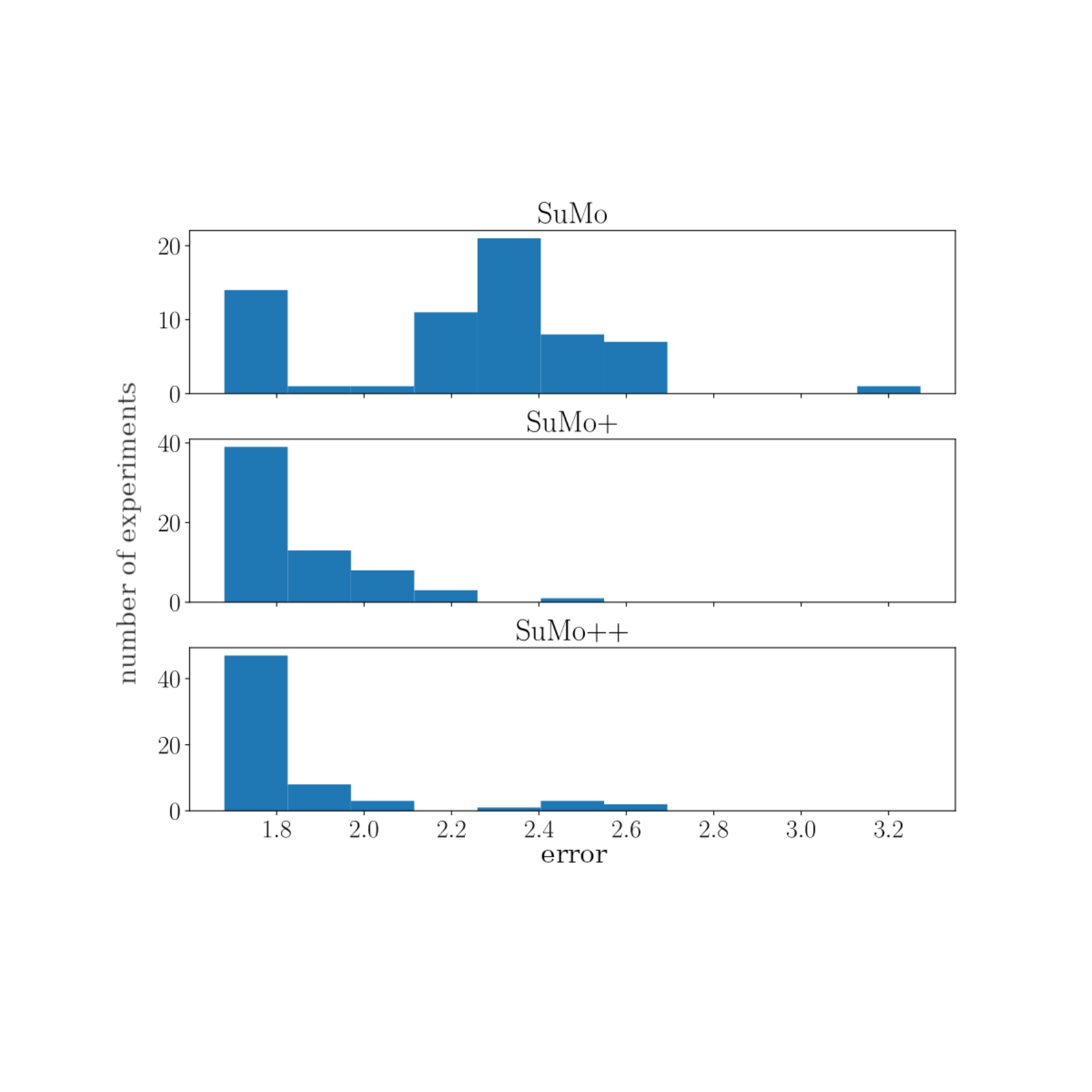}
    \caption{Analogous to Figure~\ref{fig:hists} but after convergence (2048 steps). SuMo did not improve over the 512 training steps. SuMo+ improved significantly but is still inferior to SuMo++.}
    \label{fig:hists_2048}
\end{figure}

Figure~\ref{fig:hists} shows how well SuMo, SuMo+, and SuMo++ can fit the data points of Figure~\ref{fig:toy_curves}. Figure~\ref{fig:hists_2048} shows that both SuMo+ and SuMo++ benefit from further training steps until convergences after 2048 training steps. SuMo+ almost improves to the level of SuMo++ -- while SuMo shows no significant improvement.

\section{Tables containing the separate scores}
\label{sec:separate_scores}
The Tables~\ref{tab:tab:stats_AUPRC},~\ref{tab:stats_A},~\ref{tab:stats_Accuracy},~\ref{tab:stats_Balanced accuracy},~\ref{tab:stats_F1},~\ref{tab:stats_F2},~\ref{tab:stats_F_0.5},~\ref{tab:stats_Precision},~\ref{tab:stats_Sensitivity},~\ref{tab:stats_Specificity},~\ref{tab:stats_Youden} show the separate integrated classifier metrics, and Figure~\ref{tab:stats_iibs} shows the inverted classical Brier score; we invert it so that for all of these metrics, we have that the large the better. Finally, Table~\ref{tab:stats_Concordance} also presents the concordance index based on the mean survival time, and Table~\ref{tab:stats_Min} presents the minimum of all of these scores for each model and dataset.

\begin{table*}[ht]
\centering
\begin{tabular}{l||rrrrr|rr}
Model Loss-function & COVID-19 & NKI & BGSG2 & Recur & Lymph & Clocks & California \\
\hline \hline
Kaplan-Meier & 0.56 & \textbf{0.56} & \textbf{0.65} & 0.76 & \textbf{0.54} & 0.73 & 0.74 \\
Weibull & 0.61 & 0.42 & 0.50 & 0.85 & 0.31 & $\emptyset$ & 0.76 \\
Log-logistic & 0.58 & 0.42 & 0.49 & 0.85 & 0.34 & $\emptyset$ & 0.76 \\
Log-normal & 0.58 & 0.42 & 0.50 & 0.85 & 0.35 & $\emptyset$ & 0.76 \\
Cox-piecewise & 0.62 & 0.41 & 0.49 & 0.85 & 0.32 & $\emptyset$ & 0.76 \\
Cox-spline & 0.61 & 0.42 & 0.49 & 0.85 & 0.31 & $\emptyset$ & 0.76 \\
\hline
CoxNN SuMo & 0.59 & 0.46 & 0.49 & 0.84 & 0.35 & $\emptyset$ & 0.76 \\
CoxNN BCE & 0.61 & 0.49 & 0.49 & 0.85 & 0.34 & $\emptyset$ & 0.76 \\
\cOx NN SuMo & 0.60 & 0.46 & 0.48 & 0.85 & 0.35 & $\emptyset$ & 0.75 \\
\cOx NN BCE & 0.59 & 0.40 & 0.51 & 0.85 & 0.35 & $\emptyset$ & 0.70 \\
CoxDeepNN SuMo & \textbf{0.64} & 0.38 & 0.47 & 0.84 & 0.36 & 0.83 & 0.77 \\
CoxDeepNN BCE & 0.55 & 0.47 & 0.45 & \textbf{0.86} & 0.32 & 0.91 & 0.78 \\
\hline
SuMo SuMo & 0.55 & 0.40 & 0.37 & \textbf{0.86} & 0.31 & \textbf{0.99} & 0.84 \\
SuMo BCE & 0.53 & 0.50 & 0.47 & \textbf{0.86} & 0.28 & 0.96 & 0.83 \\
SuMo+ SuMo & 0.61 & 0.37 & 0.41 & 0.84 & 0.33 & \textbf{0.99} & \textbf{0.85} \\
SuMo+ BCE & 0.58 & 0.44 & 0.49 & 0.84 & 0.29 & \textbf{0.99} & 0.82 \\
SuMo++ SuMo & 0.56 & 0.44 & 0.46 & 0.85 & 0.31 & 0.98 & \textbf{0.85} \\
SuMo++ BCE & 0.55 & 0.53 & 0.55 & 0.83 & 0.37 & 0.94 & 0.83 \\
\end{tabular}

\caption{
The over time integrated AUPRC for each model and dataset.
}
\label{tab:tab:stats_AUPRC}
\end{table*}
\begin{table*}[ht]
\centering
\begin{tabular}{l||rrrrr|rr}
Model Loss-function & COVID-19 & NKI & BGSG2 & Recur & Lymph & Clocks & California \\
\hline \hline
Kaplan-Meier & 0.50 & 0.55 & 0.50 & 0.50 & 0.55 & 0.50 & 0.50 \\
Weibull & \textbf{0.95} & 0.82 & 0.73 & \textbf{0.91} & 0.72 & $\emptyset$ & 0.86 \\
Log-logistic & \textbf{0.95} & 0.82 & 0.72 & \textbf{0.91} & 0.74 & $\emptyset$ & 0.88 \\
Log-normal & \textbf{0.95} & 0.82 & 0.72 & \textbf{0.91} & 0.75 & $\emptyset$ & 0.87 \\
Cox-piecewise & \textbf{0.95} & 0.81 & 0.73 & \textbf{0.91} & 0.72 & $\emptyset$ & 0.86 \\
Cox-spline & \textbf{0.95} & 0.81 & 0.73 & \textbf{0.91} & 0.72 & $\emptyset$ & 0.86 \\
\hline
CoxNN SuMo & \textbf{0.95} & 0.83 & 0.70 & 0.90 & 0.75 & $\emptyset$ & 0.87 \\
CoxNN BCE & \textbf{0.95} & 0.83 & 0.72 & 0.90 & 0.75 & $\emptyset$ & 0.86 \\
\cOx NN SuMo & 0.94 & 0.81 & 0.71 & \textbf{0.91} & 0.76 & $\emptyset$ & 0.81 \\
\cOx NN BCE & \textbf{0.95} & 0.82 & 0.71 & \textbf{0.91} & 0.75 & $\emptyset$ & 0.81 \\
CoxDeepNN SuMo & 0.94 & 0.81 & \textbf{0.74} & 0.90 & \textbf{0.77} & 0.78 & 0.87 \\
CoxDeepNN BCE & 0.94 & 0.82 & 0.70 & \textbf{0.91} & 0.75 & 0.96 & 0.87 \\
\hline
SuMo SuMo & 0.94 & 0.78 & 0.63 & \textbf{0.91} & 0.75 & \textbf{0.99} & 0.92 \\
SuMo BCE & 0.93 & 0.83 & 0.72 & \textbf{0.91} & 0.72 & 0.98 & 0.93 \\
SuMo+ SuMo & 0.92 & 0.77 & 0.66 & 0.90 & 0.72 & \textbf{0.99} & \textbf{0.95} \\
SuMo+ BCE & 0.94 & 0.80 & 0.71 & \textbf{0.91} & 0.73 & \textbf{0.99} & 0.91 \\
SuMo++ SuMo & 0.91 & 0.80 & 0.72 & \textbf{0.91} & 0.74 & \textbf{0.99} & 0.94 \\
SuMo++ BCE & 0.88 & \textbf{0.86} & \textbf{0.74} & 0.89 & 0.74 & 0.97 & 0.93 \\
\end{tabular}

\caption{
The over time integrated AUROC for each model and dataset.
}
\label{tab:stats_A}
\end{table*}
\begin{table*}[ht]
\centering
\begin{tabular}{l||rrrrr|rr}
Model Loss-function & COVID-19 & NKI & BGSG2 & Recur & Lymph & Clocks & California \\
\hline \hline
Kaplan-Meier & 0.79 & 0.69 & 0.63 & 0.57 & 0.74 & 0.63 & 0.69 \\
Weibull & 0.90 & 0.76 & 0.67 & 0.77 & 0.76 & $\emptyset$ & 0.80 \\
Log-logistic & 0.90 & 0.77 & 0.67 & 0.78 & 0.76 & $\emptyset$ & 0.83 \\
Log-normal & 0.90 & 0.77 & 0.67 & 0.78 & 0.76 & $\emptyset$ & 0.82 \\
Cox-piecewise & 0.90 & 0.76 & 0.67 & 0.77 & 0.76 & $\emptyset$ & 0.80 \\
Cox-spline & 0.90 & 0.76 & 0.67 & 0.77 & 0.76 & $\emptyset$ & 0.80 \\
\hline
CoxNN SuMo & 0.90 & 0.69 & 0.63 & 0.79 & 0.70 & $\emptyset$ & 0.78 \\
CoxNN BCE & 0.86 & 0.60 & 0.66 & 0.74 & 0.67 & $\emptyset$ & 0.70 \\
\cOx NN SuMo & 0.91 & 0.72 & 0.65 & 0.82 & 0.70 & $\emptyset$ & 0.78 \\
\cOx NN BCE & 0.84 & 0.74 & 0.63 & 0.81 & 0.70 & $\emptyset$ & 0.73 \\
CoxDeepNN SuMo & 0.91 & 0.73 & 0.64 & 0.81 & \textbf{0.78} & 0.86 & 0.78 \\
CoxDeepNN BCE & 0.87 & 0.66 & 0.64 & 0.80 & 0.72 & 0.87 & 0.71 \\
\hline
SuMo SuMo & \textbf{0.92} & 0.77 & 0.66 & 0.82 & \textbf{0.78} & \textbf{0.98} & \textbf{0.89} \\
SuMo BCE & 0.90 & 0.72 & 0.71 & \textbf{0.83} & 0.72 & 0.93 & 0.87 \\
SuMo+ SuMo & 0.91 & 0.76 & 0.67 & 0.81 & 0.75 & \textbf{0.98} & \textbf{0.89} \\
SuMo+ BCE & 0.90 & 0.75 & 0.71 & 0.80 & 0.74 & 0.94 & 0.88 \\
SuMo++ SuMo & 0.90 & 0.76 & 0.65 & 0.82 & 0.75 & 0.97 & \textbf{0.89} \\
SuMo++ BCE & 0.90 & \textbf{0.78} & \textbf{0.72} & 0.81 & 0.76 & 0.93 & 0.87 \\
\end{tabular}

\caption{
The over time integrated accuracy score for each model and dataset.
}
\label{tab:stats_Accuracy}
\end{table*}
\begin{table*}[ht]
\centering
\begin{tabular}{l||rrrrr|rr}
Model Loss-function & COVID-19 & NKI & BGSG2 & Recur & Lymph & Clocks & California \\
\hline \hline
Kaplan-Meier & 0.50 & 0.50 & 0.50 & 0.50 & 0.50 & 0.50 & 0.50 \\
Weibull & 0.69 & 0.56 & 0.55 & 0.72 & 0.54 & $\emptyset$ & 0.64 \\
Log-logistic & 0.70 & 0.58 & 0.55 & 0.74 & 0.54 & $\emptyset$ & 0.68 \\
Log-normal & 0.70 & 0.57 & 0.55 & 0.75 & 0.54 & $\emptyset$ & 0.67 \\
Cox-piecewise & 0.69 & 0.57 & 0.55 & 0.73 & 0.54 & $\emptyset$ & 0.64 \\
Cox-spline & 0.69 & 0.57 & 0.55 & 0.72 & 0.54 & $\emptyset$ & 0.64 \\
\hline
CoxNN SuMo & 0.74 & 0.60 & 0.54 & 0.75 & 0.55 & $\emptyset$ & 0.64 \\
CoxNN BCE & 0.81 & 0.59 & 0.55 & 0.69 & 0.55 & $\emptyset$ & 0.60 \\
\cOx NN SuMo & 0.79 & 0.62 & 0.56 & 0.78 & 0.57 & $\emptyset$ & 0.62 \\
\cOx NN BCE & 0.80 & 0.62 & 0.55 & 0.78 & 0.55 & $\emptyset$ & 0.59 \\
CoxDeepNN SuMo & 0.78 & 0.59 & \textbf{0.59} & 0.77 & 0.56 & 0.72 & 0.65 \\
CoxDeepNN BCE & 0.82 & 0.65 & 0.53 & 0.77 & 0.57 & 0.81 & 0.61 \\
\hline
SuMo SuMo & 0.77 & 0.61 & 0.55 & 0.79 & 0.58 & \textbf{0.97} & 0.78 \\
SuMo BCE & 0.83 & 0.69 & \textbf{0.59} & \textbf{0.80} & 0.55 & 0.91 & 0.76 \\
SuMo+ SuMo & 0.77 & 0.57 & 0.56 & 0.79 & 0.57 & 0.96 & \textbf{0.79} \\
SuMo+ BCE & \textbf{0.86} & 0.66 & \textbf{0.59} & 0.78 & 0.57 & 0.90 & 0.77 \\
SuMo++ SuMo & 0.80 & 0.64 & 0.57 & 0.79 & 0.57 & 0.95 & 0.78 \\
SuMo++ BCE & 0.79 & \textbf{0.70} & \textbf{0.59} & 0.78 & \textbf{0.60} & 0.92 & 0.77 \\
\end{tabular}

\caption{
The over time integrated balanced accuracy score for each model and dataset.
}
\label{tab:stats_Balanced accuracy}
\end{table*}
\begin{table*}[ht]
\centering
\begin{tabular}{l||rrrrr|rr}
Model Loss-function & COVID-19 & NKI & BGSG2 & Recur & Lymph & Clocks & California \\
\hline \hline
Kaplan-Meier & 0.50 & 0.50 & 0.50 & 0.50 & 0.50 & 0.50 & 0.50 \\
Weibull & 0.93 & 0.76 & \textbf{0.70} & 0.82 & 0.67 & $\emptyset$ & 0.81 \\
Log-logistic & 0.93 & 0.76 & 0.69 & 0.82 & 0.68 & $\emptyset$ & 0.82 \\
Log-normal & 0.93 & 0.76 & 0.69 & 0.82 & 0.69 & $\emptyset$ & 0.81 \\
Cox-piecewise & 0.93 & 0.75 & \textbf{0.70} & 0.82 & 0.67 & $\emptyset$ & 0.81 \\
Cox-spline & 0.93 & 0.75 & \textbf{0.70} & 0.82 & 0.67 & $\emptyset$ & 0.81 \\
\hline
CoxNN SuMo & 0.93 & 0.77 & 0.67 & 0.82 & 0.70 & $\emptyset$ & 0.82 \\
CoxNN BCE & \textbf{0.94} & 0.77 & 0.69 & 0.82 & 0.69 & $\emptyset$ & 0.81 \\
\cOx NN SuMo & 0.93 & 0.75 & 0.68 & 0.82 & \textbf{0.71} & $\emptyset$ & 0.78 \\
\cOx NN BCE & 0.93 & 0.76 & 0.67 & \textbf{0.83} & 0.70 & $\emptyset$ & 0.75 \\
CoxDeepNN SuMo & 0.93 & 0.75 & 0.69 & 0.82 & 0.70 & 0.88 & 0.83 \\
CoxDeepNN BCE & 0.92 & 0.76 & 0.66 & 0.82 & 0.69 & 0.90 & 0.82 \\
\hline
SuMo SuMo & 0.92 & 0.71 & 0.60 & 0.82 & 0.68 & 0.97 & 0.87 \\
SuMo BCE & 0.92 & 0.75 & 0.69 & 0.82 & 0.65 & 0.92 & 0.86 \\
SuMo+ SuMo & 0.91 & 0.68 & 0.63 & 0.81 & 0.66 & \textbf{0.98} & \textbf{0.88} \\
SuMo+ BCE & 0.93 & 0.72 & 0.68 & 0.81 & 0.67 & 0.94 & 0.86 \\
SuMo++ SuMo & 0.90 & 0.71 & 0.67 & 0.81 & 0.67 & 0.96 & \textbf{0.88} \\
SuMo++ BCE & 0.87 & \textbf{0.79} & 0.69 & 0.80 & 0.69 & 0.91 & 0.86 \\
\end{tabular}

\caption{The concordance index based on the dataset's restricted mean survival time for each model and dataset.}
\label{tab:stats_Concordance}
\end{table*}
\begin{table*}[ht]
\centering
\begin{tabular}{l||rrrrr|rr}
Model Loss-function & COVID-19 & NKI & BGSG2 & Recur & Lymph & Clocks & California \\
\hline \hline
Kaplan-Meier & 0.12 & 0.21 & 0.30 & 0.53 & 0.17 & 0.48 & 0.48 \\
Weibull & 0.46 & 0.29 & 0.36 & 0.73 & 0.24 & $\emptyset$ & 0.61 \\
Log-logistic & 0.48 & 0.31 & 0.37 & 0.74 & 0.25 & $\emptyset$ & 0.65 \\
Log-normal & 0.48 & 0.31 & 0.37 & 0.75 & 0.25 & $\emptyset$ & 0.64 \\
Cox-piecewise & 0.46 & 0.30 & 0.36 & 0.73 & 0.24 & $\emptyset$ & 0.61 \\
Cox-spline & 0.46 & 0.30 & 0.36 & 0.73 & 0.24 & $\emptyset$ & 0.61 \\
\hline
CoxNN SuMo & 0.52 & 0.37 & 0.39 & 0.76 & 0.30 & $\emptyset$ & 0.62 \\
CoxNN BCE & 0.52 & 0.39 & 0.36 & 0.74 & 0.29 & $\emptyset$ & 0.58 \\
\cOx NN SuMo & 0.58 & 0.39 & 0.40 & 0.77 & 0.31 & $\emptyset$ & 0.60 \\
\cOx NN BCE & 0.49 & 0.39 & 0.39 & 0.78 & 0.28 & $\emptyset$ & 0.58 \\
CoxDeepNN SuMo & 0.56 & 0.35 & \textbf{0.46} & 0.78 & 0.27 & 0.55 & 0.64 \\
CoxDeepNN BCE & 0.54 & 0.43 & 0.35 & 0.78 & 0.30 & 0.82 & 0.58 \\
\hline
SuMo SuMo & 0.58 & 0.38 & 0.39 & 0.78 & 0.31 & \textbf{0.97} & 0.75 \\
SuMo BCE & 0.59 & 0.49 & 0.39 & \textbf{0.81} & 0.28 & 0.88 & 0.72 \\
SuMo+ SuMo & 0.55 & 0.30 & 0.40 & 0.77 & 0.31 & 0.96 & \textbf{0.76} \\
SuMo+ BCE & \textbf{0.61} & 0.45 & 0.39 & 0.79 & 0.29 & 0.89 & 0.73 \\
SuMo++ SuMo & 0.57 & 0.42 & 0.44 & 0.78 & 0.29 & 0.94 & 0.75 \\
SuMo++ BCE & 0.57 & \textbf{0.50} & 0.40 & 0.80 & \textbf{0.33} & 0.88 & 0.72 \\
\end{tabular}

\caption{
The over time integrated $F_{1}$ score for each model and dataset.
}
\label{tab:stats_F1}
\end{table*}
\begin{table*}[ht]
\centering
\begin{tabular}{l||rrrrr|rr}
Model Loss-function & COVID-19 & NKI & BGSG2 & Recur & Lymph & Clocks & California \\
\hline \hline
Kaplan-Meier & 0.12 & 0.21 & 0.30 & 0.52 & 0.17 & 0.49 & 0.48 \\
Weibull & 0.44 & 0.27 & 0.36 & 0.74 & 0.24 & $\emptyset$ & 0.62 \\
Log-logistic & 0.46 & 0.29 & 0.37 & 0.76 & 0.24 & $\emptyset$ & 0.66 \\
Log-normal & 0.47 & 0.29 & 0.36 & 0.76 & 0.25 & $\emptyset$ & 0.64 \\
Cox-piecewise & 0.44 & 0.28 & 0.36 & 0.74 & 0.24 & $\emptyset$ & 0.62 \\
Cox-spline & 0.44 & 0.28 & 0.36 & 0.74 & 0.24 & $\emptyset$ & 0.62 \\
\hline
CoxNN SuMo & 0.53 & 0.44 & 0.43 & 0.78 & 0.35 & $\emptyset$ & 0.65 \\
CoxNN BCE & 0.63 & 0.51 & 0.35 & 0.81 & 0.36 & $\emptyset$ & 0.62 \\
\cOx NN SuMo & 0.62 & 0.45 & 0.44 & 0.78 & \textbf{0.37} & $\emptyset$ & 0.62 \\
\cOx NN BCE & 0.62 & 0.43 & 0.42 & 0.82 & 0.33 & $\emptyset$ & 0.61 \\
CoxDeepNN SuMo & 0.60 & 0.38 & \textbf{0.54} & 0.80 & 0.26 & 0.55 & 0.70 \\
CoxDeepNN BCE & 0.65 & 0.55 & 0.36 & 0.83 & 0.33 & 0.88 & 0.62 \\
\hline
SuMo SuMo & 0.58 & 0.39 & 0.43 & 0.80 & 0.33 & \textbf{0.96} & 0.74 \\
SuMo BCE & 0.67 & \textbf{0.57} & 0.37 & \textbf{0.85} & 0.30 & 0.92 & 0.73 \\
SuMo+ SuMo & 0.58 & 0.28 & 0.41 & 0.76 & 0.34 & 0.95 & \textbf{0.76} \\
SuMo+ BCE & \textbf{0.70} & 0.50 & 0.36 & 0.83 & 0.31 & 0.93 & 0.75 \\
SuMo++ SuMo & 0.63 & 0.46 & 0.50 & 0.79 & 0.31 & 0.94 & 0.75 \\
SuMo++ BCE & 0.61 & 0.55 & 0.37 & 0.84 & 0.36 & 0.92 & 0.73 \\
\end{tabular}

\caption{
The over time integrated $F_{2}$ score for each model and dataset.
}
\label{tab:stats_F2}
\end{table*}
\begin{table*}[ht]
\centering
\begin{tabular}{l||rrrrr|rr}
Model Loss-function & COVID-19 & NKI & BGSG2 & Recur & Lymph & Clocks & California \\
\hline \hline
Kaplan-Meier & 0.12 & 0.21 & 0.30 & 0.53 & 0.18 & 0.47 & 0.48 \\
Weibull & 0.48 & 0.32 & 0.36 & 0.71 & 0.25 & $\emptyset$ & 0.60 \\
Log-logistic & 0.50 & 0.33 & 0.37 & 0.73 & 0.25 & $\emptyset$ & 0.65 \\
Log-normal & 0.50 & 0.34 & 0.37 & 0.73 & 0.26 & $\emptyset$ & 0.64 \\
Cox-piecewise & 0.49 & 0.32 & 0.36 & 0.72 & 0.25 & $\emptyset$ & 0.60 \\
Cox-spline & 0.49 & 0.32 & 0.36 & 0.72 & 0.25 & $\emptyset$ & 0.60 \\
\hline
CoxNN SuMo & 0.50 & 0.33 & 0.36 & 0.74 & 0.26 & $\emptyset$ & 0.59 \\
CoxNN BCE & 0.45 & 0.31 & 0.36 & 0.69 & 0.25 & $\emptyset$ & 0.55 \\
\cOx NN SuMo & 0.55 & 0.35 & 0.37 & 0.77 & 0.27 & $\emptyset$ & 0.58 \\
\cOx NN BCE & 0.41 & 0.36 & 0.37 & 0.75 & 0.25 & $\emptyset$ & 0.55 \\
CoxDeepNN SuMo & 0.54 & 0.33 & 0.41 & 0.76 & 0.28 & 0.55 & 0.60 \\
CoxDeepNN BCE & 0.47 & 0.36 & 0.35 & 0.74 & 0.27 & 0.77 & 0.56 \\
\hline
SuMo SuMo & \textbf{0.58} & 0.37 & 0.37 & 0.77 & 0.29 & \textbf{0.97} & \textbf{0.75} \\
SuMo BCE & 0.54 & 0.44 & 0.42 & \textbf{0.78} & 0.26 & 0.85 & 0.71 \\
SuMo+ SuMo & 0.53 & 0.32 & 0.38 & \textbf{0.78} & 0.29 & \textbf{0.97} & \textbf{0.75} \\
SuMo+ BCE & 0.54 & 0.41 & \textbf{0.43} & 0.75 & 0.27 & 0.86 & 0.72 \\
SuMo++ SuMo & 0.53 & 0.39 & 0.40 & 0.77 & 0.28 & 0.94 & \textbf{0.75} \\
SuMo++ BCE & 0.55 & \textbf{0.47} & \textbf{0.43} & 0.76 & \textbf{0.31} & 0.85 & 0.71 \\
\end{tabular}

\caption{
The over time integrated $F_{0.5}$ score for each model and dataset.
}
\label{tab:stats_F_0.5}
\end{table*}
\begin{table*}[ht]
\centering
\begin{tabular}{l||rrrrr|rr}
Model Loss-function & COVID-19 & NKI & BGSG2 & Recur & Lymph & Clocks & California \\
\hline \hline
Kaplan-Meier & 0.00 & 0.00 & 0.00 & 0.00 & 0.00 & 0.00 & 0.00 \\
Weibull & 0.38 & 0.13 & 0.10 & 0.44 & 0.08 & $\emptyset$ & 0.29 \\
Log-logistic & 0.40 & 0.16 & 0.10 & 0.49 & 0.08 & $\emptyset$ & 0.36 \\
Log-normal & 0.41 & 0.15 & 0.10 & 0.49 & 0.09 & $\emptyset$ & 0.33 \\
Cox-piecewise & 0.38 & 0.14 & 0.10 & 0.45 & 0.08 & $\emptyset$ & 0.28 \\
Cox-spline & 0.38 & 0.13 & 0.10 & 0.45 & 0.08 & $\emptyset$ & 0.28 \\
\hline
CoxNN SuMo & 0.48 & 0.21 & 0.09 & 0.50 & 0.12 & $\emptyset$ & 0.29 \\
CoxNN BCE & 0.41 & 0.22 & 0.09 & 0.39 & 0.12 & $\emptyset$ & 0.20 \\
\cOx NN SuMo & 0.53 & 0.24 & 0.11 & 0.57 & 0.14 & $\emptyset$ & 0.24 \\
\cOx NN BCE & 0.37 & 0.25 & 0.09 & 0.56 & 0.11 & $\emptyset$ & 0.17 \\
CoxDeepNN SuMo & 0.52 & 0.19 & \textbf{0.19} & 0.55 & 0.13 & 0.44 & 0.31 \\
CoxDeepNN BCE & 0.43 & 0.32 & 0.07 & 0.54 & 0.15 & 0.63 & 0.22 \\
\hline
SuMo SuMo & \textbf{0.54} & 0.25 & 0.10 & 0.58 & 0.16 & \textbf{0.94} & 0.56 \\
SuMo BCE & 0.51 & 0.41 & 0.18 & \textbf{0.60} & 0.13 & 0.82 & 0.51 \\
SuMo+ SuMo & 0.52 & 0.17 & 0.12 & 0.57 & 0.15 & 0.92 & \textbf{0.59} \\
SuMo+ BCE & 0.51 & 0.35 & 0.18 & 0.56 & 0.14 & 0.81 & 0.53 \\
SuMo++ SuMo & 0.50 & 0.30 & 0.15 & 0.58 & 0.14 & 0.90 & 0.57 \\
SuMo++ BCE & \textbf{0.54} & \textbf{0.44} & \textbf{0.19} & 0.57 & \textbf{0.20} & 0.83 & 0.54 \\
\end{tabular}

\caption{
The minimum of all over time integrated scores for each model and dataset.
}
\label{tab:stats_Min}
\end{table*}
\begin{table*}[ht]
\centering
\begin{tabular}{l||rrrrr|rr}
Model Loss-function & COVID-19 & NKI & BGSG2 & Recur & Lymph & Clocks & California \\
\hline \hline
Kaplan-Meier & 0.12 & 0.21 & 0.30 & 0.53 & 0.19 & 0.47 & 0.48 \\
Weibull & 0.50 & 0.33 & 0.37 & 0.71 & 0.25 & $\emptyset$ & 0.60 \\
Log-logistic & 0.51 & 0.35 & 0.37 & 0.73 & 0.26 & $\emptyset$ & 0.65 \\
Log-normal & 0.51 & 0.36 & 0.37 & 0.73 & 0.26 & $\emptyset$ & 0.64 \\
Cox-piecewise & 0.50 & 0.33 & 0.37 & 0.71 & 0.25 & $\emptyset$ & 0.60 \\
Cox-spline & 0.51 & 0.33 & 0.37 & 0.71 & 0.25 & $\emptyset$ & 0.60 \\
\hline
CoxNN SuMo & 0.49 & 0.30 & 0.34 & 0.72 & 0.24 & $\emptyset$ & 0.58 \\
CoxNN BCE & 0.41 & 0.27 & 0.36 & 0.65 & 0.23 & $\emptyset$ & 0.54 \\
\cOx NN SuMo & 0.53 & 0.32 & 0.36 & 0.77 & 0.25 & $\emptyset$ & 0.57 \\
\cOx NN BCE & 0.37 & 0.34 & 0.35 & 0.73 & 0.23 & $\emptyset$ & 0.54 \\
CoxDeepNN SuMo & 0.52 & 0.32 & 0.38 & 0.75 & 0.29 & 0.56 & 0.57 \\
CoxDeepNN BCE & 0.43 & 0.32 & 0.35 & 0.72 & 0.26 & 0.74 & 0.55 \\
\hline
SuMo SuMo & \textbf{0.58} & 0.36 & 0.36 & 0.76 & 0.29 & 0.97 & \textbf{0.76} \\
SuMo BCE & 0.51 & 0.41 & 0.45 & 0.76 & 0.25 & 0.84 & 0.70 \\
SuMo+ SuMo & 0.52 & 0.33 & 0.38 & \textbf{0.78} & 0.28 & \textbf{0.98} & 0.75 \\
SuMo+ BCE & 0.51 & 0.39 & 0.45 & 0.73 & 0.26 & 0.85 & 0.71 \\
SuMo++ SuMo & 0.50 & 0.37 & 0.37 & 0.77 & 0.27 & 0.94 & 0.75 \\
SuMo++ BCE & 0.54 & \textbf{0.45} & \textbf{0.46} & 0.74 & \textbf{0.30} & 0.83 & 0.70 \\
\end{tabular}

\caption{
The over time integrated precision score for each model and dataset.
}
\label{tab:stats_Precision}
\end{table*}
\begin{table*}[ht]
\centering
\begin{tabular}{l||rrrrr|rr}
Model Loss-function & COVID-19 & NKI & BGSG2 & Recur & Lymph & Clocks & California \\
\hline \hline
Kaplan-Meier & 0.12 & 0.21 & 0.29 & 0.52 & 0.16 & 0.50 & 0.48 \\
Weibull & 0.43 & 0.26 & 0.35 & 0.75 & 0.23 & $\emptyset$ & 0.62 \\
Log-logistic & 0.45 & 0.28 & 0.37 & 0.77 & 0.24 & $\emptyset$ & 0.66 \\
Log-normal & 0.46 & 0.27 & 0.36 & 0.77 & 0.24 & $\emptyset$ & 0.64 \\
Cox-piecewise & 0.43 & 0.27 & 0.36 & 0.75 & 0.24 & $\emptyset$ & 0.63 \\
Cox-spline & 0.43 & 0.27 & 0.37 & 0.75 & 0.24 & $\emptyset$ & 0.62 \\
\hline
CoxNN SuMo & 0.55 & 0.50 & 0.47 & 0.80 & 0.40 & $\emptyset$ & 0.69 \\
CoxNN BCE & 0.75 & 0.65 & 0.35 & 0.86 & \textbf{0.42} & $\emptyset$ & 0.68 \\
\cOx NN SuMo & 0.65 & 0.50 & 0.47 & 0.78 & \textbf{0.42} & $\emptyset$ & 0.63 \\
\cOx NN BCE & 0.76 & 0.46 & 0.45 & 0.84 & 0.37 & $\emptyset$ & 0.65 \\
CoxDeepNN SuMo & 0.63 & 0.41 & \textbf{0.61} & 0.81 & 0.26 & 0.55 & \textbf{0.76} \\
CoxDeepNN BCE & 0.77 & \textbf{0.69} & 0.36 & 0.86 & 0.36 & 0.93 & 0.69 \\
\hline
SuMo SuMo & 0.58 & 0.40 & 0.46 & 0.81 & 0.34 & 0.96 & 0.74 \\
SuMo BCE & 0.74 & 0.67 & 0.36 & \textbf{0.88} & 0.33 & 0.95 & 0.74 \\
SuMo+ SuMo & 0.60 & 0.27 & 0.43 & 0.75 & 0.37 & 0.94 & \textbf{0.76} \\
SuMo+ BCE & \textbf{0.81} & 0.55 & 0.35 & 0.86 & 0.34 & \textbf{0.97} & \textbf{0.76} \\
SuMo++ SuMo & 0.69 & 0.49 & 0.55 & 0.80 & 0.33 & 0.94 & 0.75 \\
SuMo++ BCE & 0.65 & 0.60 & 0.36 & 0.87 & 0.39 & 0.96 & 0.74 \\
\end{tabular}

\caption{
The over time integrated sensitivity score for each model and dataset.
}
\label{tab:stats_Sensitivity}
\end{table*}
\begin{table*}[ht]
\centering
\begin{tabular}{l||rrrrr|rr}
Model Loss-function & COVID-19 & NKI & BGSG2 & Recur & Lymph & Clocks & California \\
\hline \hline
Kaplan-Meier & 0.88 & 0.79 & 0.71 & 0.48 & 0.84 & 0.50 & 0.52 \\
Weibull & 0.95 & 0.87 & 0.74 & 0.69 & 0.84 & $\emptyset$ & 0.67 \\
Log-logistic & 0.95 & 0.87 & 0.74 & 0.72 & 0.85 & $\emptyset$ & 0.70 \\
Log-normal & 0.95 & \textbf{0.88} & 0.74 & 0.72 & 0.85 & $\emptyset$ & 0.69 \\
Cox-piecewise & 0.95 & 0.86 & 0.73 & 0.71 & 0.84 & $\emptyset$ & 0.65 \\
Cox-spline & 0.95 & 0.86 & 0.73 & 0.70 & 0.84 & $\emptyset$ & 0.66 \\
\hline
CoxNN SuMo & 0.93 & 0.69 & 0.62 & 0.70 & 0.71 & $\emptyset$ & 0.60 \\
CoxNN BCE & 0.87 & 0.54 & 0.74 & 0.52 & 0.69 & $\emptyset$ & 0.52 \\
\cOx NN SuMo & 0.93 & 0.73 & 0.65 & 0.78 & 0.71 & $\emptyset$ & 0.60 \\
\cOx NN BCE & 0.84 & 0.77 & 0.65 & 0.72 & 0.73 & $\emptyset$ & 0.52 \\
CoxDeepNN SuMo & 0.93 & 0.77 & 0.58 & 0.73 & \textbf{0.87} & 0.89 & 0.54 \\
CoxDeepNN BCE & 0.88 & 0.61 & 0.71 & 0.68 & 0.77 & 0.70 & 0.53 \\
\hline
SuMo SuMo & \textbf{0.96} & 0.83 & 0.64 & 0.76 & 0.81 & \textbf{0.98} & \textbf{0.83} \\
SuMo BCE & 0.92 & 0.72 & 0.82 & 0.72 & 0.78 & 0.86 & 0.77 \\
SuMo+ SuMo & 0.94 & 0.86 & 0.69 & \textbf{0.82} & 0.78 & \textbf{0.98} & \textbf{0.83} \\
SuMo+ BCE & 0.90 & 0.77 & \textbf{0.83} & 0.70 & 0.79 & 0.83 & 0.78 \\
SuMo++ SuMo & 0.92 & 0.79 & 0.60 & 0.78 & 0.80 & 0.96 & 0.82 \\
SuMo++ BCE & 0.93 & 0.81 & \textbf{0.83} & 0.70 & 0.80 & 0.88 & 0.79 \\
\end{tabular}

\caption{
The over time integrated specificity score for each model and dataset.
}
\label{tab:stats_Specificity}
\end{table*}
\begin{table*}[ht]
\centering
\begin{tabular}{l||rrrrr|rr}
Model Loss-function & COVID-19 & NKI & BGSG2 & Recur & Lymph & Clocks & California \\
\hline \hline
Kaplan-Meier & 0.00 & 0.00 & 0.00 & 0.00 & 0.00 & 0.00 & 0.00 \\
Weibull & 0.38 & 0.13 & 0.10 & 0.44 & 0.08 & $\emptyset$ & 0.29 \\
Log-logistic & 0.40 & 0.16 & 0.10 & 0.49 & 0.08 & $\emptyset$ & 0.36 \\
Log-normal & 0.41 & 0.15 & 0.10 & 0.49 & 0.09 & $\emptyset$ & 0.33 \\
Cox-piecewise & 0.38 & 0.14 & 0.10 & 0.45 & 0.08 & $\emptyset$ & 0.28 \\
Cox-spline & 0.38 & 0.13 & 0.10 & 0.45 & 0.08 & $\emptyset$ & 0.28 \\
\hline
CoxNN SuMo & 0.48 & 0.21 & 0.09 & 0.50 & 0.12 & $\emptyset$ & 0.29 \\
CoxNN BCE & 0.62 & 0.22 & 0.09 & 0.39 & 0.12 & $\emptyset$ & 0.20 \\
\cOx NN SuMo & 0.58 & 0.24 & 0.11 & 0.57 & 0.14 & $\emptyset$ & 0.24 \\
\cOx NN BCE & 0.60 & 0.25 & 0.09 & 0.56 & 0.11 & $\emptyset$ & 0.17 \\
CoxDeepNN SuMo & 0.56 & 0.19 & \textbf{0.19} & 0.55 & 0.13 & 0.44 & 0.31 \\
CoxDeepNN BCE & 0.64 & 0.34 & 0.07 & 0.54 & 0.15 & 0.63 & 0.22 \\
\hline
SuMo SuMo & 0.54 & 0.25 & 0.10 & 0.58 & 0.16 & \textbf{0.94} & 0.56 \\
SuMo BCE & 0.65 & \textbf{0.46} & 0.18 & \textbf{0.60} & 0.13 & 0.82 & 0.51 \\
SuMo+ SuMo & 0.54 & 0.17 & 0.12 & 0.57 & 0.15 & 0.92 & \textbf{0.59} \\
SuMo+ BCE & \textbf{0.72} & 0.35 & 0.18 & 0.56 & 0.14 & 0.81 & 0.53 \\
SuMo++ SuMo & 0.61 & 0.30 & 0.15 & 0.58 & 0.14 & 0.90 & 0.57 \\
SuMo++ BCE & 0.58 & 0.44 & \textbf{0.19} & 0.57 & \textbf{0.20} & 0.83 & 0.54 \\
\end{tabular}

\caption{
The over time integrated Youden's index for each model and dataset.
}
\label{tab:stats_Youden}
\end{table*}
\begin{table*}[ht]
\centering
\begin{tabular}{l||rrrrr|rr}
Model Loss-function & COVID-19 & NKI & BGSG2 & Recur & Lymph & Clocks & California \\
\hline \hline
Kaplan-Meier & 0.90 & 0.84 & 0.81 & 0.78 & 0.86 & 0.81 & 0.85 \\
Weibull & 0.94 & \textbf{0.87} & 0.83 & 0.89 & 0.87 & $\emptyset$ & 0.91 \\
Log-logistic & 0.94 & \textbf{0.87} & \textbf{0.84} & 0.89 & 0.87 & $\emptyset$ & 0.92 \\
Log-normal & 0.94 & \textbf{0.87} & 0.83 & 0.89 & \textbf{0.88} & $\emptyset$ & 0.91 \\
Cox-piecewise & 0.94 & \textbf{0.87} & \textbf{0.84} & 0.89 & 0.87 & $\emptyset$ & 0.91 \\
Cox-spline & 0.94 & \textbf{0.87} & \textbf{0.84} & 0.89 & 0.87 & $\emptyset$ & 0.91 \\
\hline
CoxNN SuMo & \textbf{0.95} & 0.85 & 0.82 & 0.89 & 0.85 & $\emptyset$ & 0.90 \\
CoxNN BCE & 0.93 & 0.78 & 0.83 & 0.85 & 0.85 & $\emptyset$ & 0.87 \\
\cOx NN SuMo & 0.94 & 0.85 & 0.82 & \textbf{0.90} & 0.85 & $\emptyset$ & 0.89 \\
\cOx NN BCE & 0.93 & 0.86 & 0.83 & 0.89 & 0.86 & $\emptyset$ & 0.88 \\
CoxDeepNN SuMo & 0.94 & 0.85 & 0.80 & 0.89 & \textbf{0.88} & 0.88 & 0.91 \\
CoxDeepNN BCE & 0.92 & 0.80 & 0.83 & 0.89 & 0.87 & 0.93 & 0.88 \\
\hline
SuMo SuMo & 0.94 & 0.82 & 0.77 & \textbf{0.90} & 0.84 & 0.98 & 0.94 \\
SuMo BCE & 0.92 & 0.75 & 0.82 & 0.88 & 0.83 & 0.96 & 0.93 \\
SuMo+ SuMo & 0.94 & 0.81 & 0.79 & 0.89 & 0.83 & \textbf{0.99} & 0.94 \\
SuMo+ BCE & 0.92 & 0.81 & 0.81 & 0.89 & 0.85 & 0.96 & 0.93 \\
SuMo++ SuMo & 0.94 & 0.83 & 0.79 & \textbf{0.90} & 0.85 & 0.98 & \textbf{0.95} \\
SuMo++ BCE & 0.92 & 0.84 & 0.83 & 0.88 & 0.84 & 0.96 & 0.93 \\
\end{tabular}

\caption{
The over time integrated inverted mean squared error defined as $1 - \text{MSE}$ for each model and dataset, I.e., $1 - \text{IBS}$ where IBS is the integrated Brier score.
}
\label{tab:stats_iibs}
\end{table*}

\FloatBarrier
\section{Comparison of non-integrated balanced accuracy and F1 scores using plots}
\label{sec:plots_non_integrated}
\begin{figure*}
    \centering
    \includegraphics[trim={35mm 5mm 35mm 5mm}, clip, width=1\textwidth]{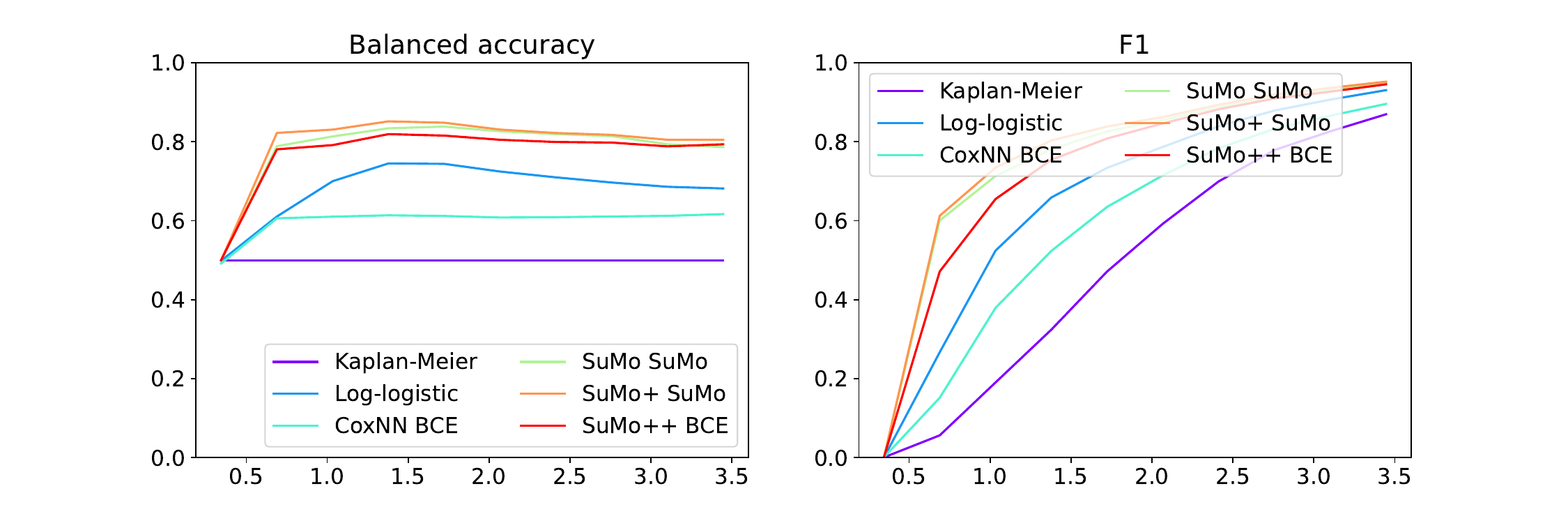}
    \caption{The balanced accuracy and the $F_1$ score over time for the California dataset.}
    \label{fig:ot_california}
\end{figure*}

\begin{figure*}
    \centering
    \includegraphics[trim={35mm 5mm 35mm 5mm}, clip, width=1\textwidth]{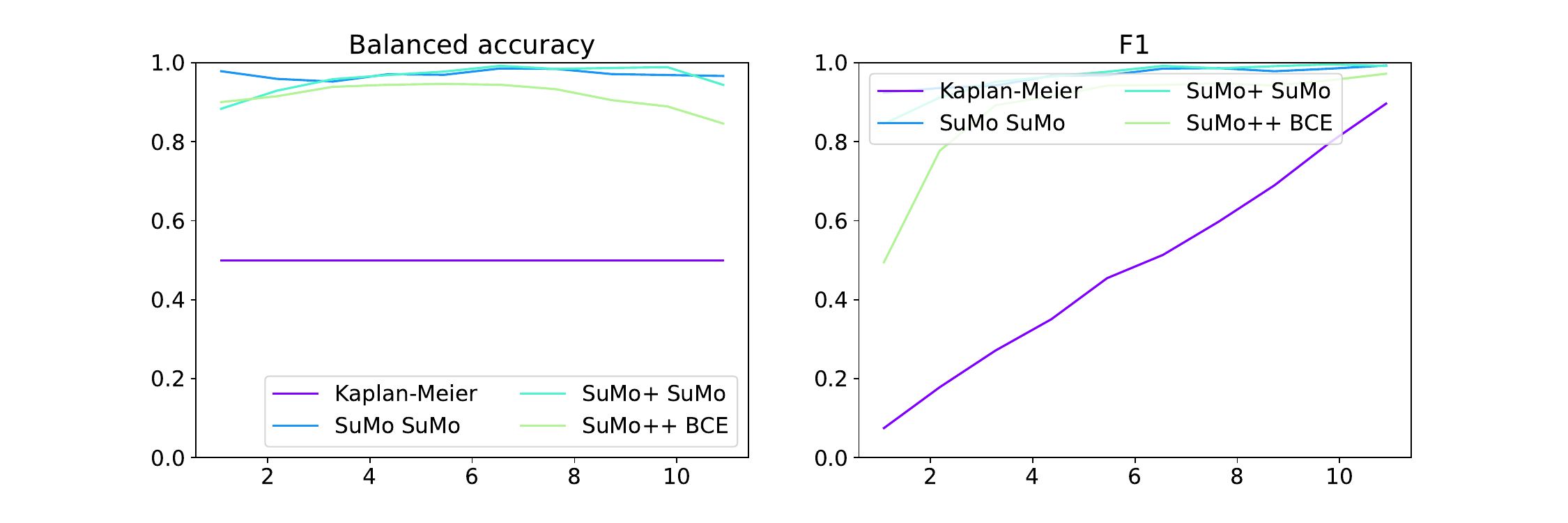}
    \caption{The balanced accuracy and the $F_1$ score over time for the Clocks dataset.}
    \label{fig:ot_clocks}
\end{figure*}

\begin{figure*}
    \centering
    \includegraphics[trim={35mm 5mm 35mm 5mm}, clip, width=1\textwidth]{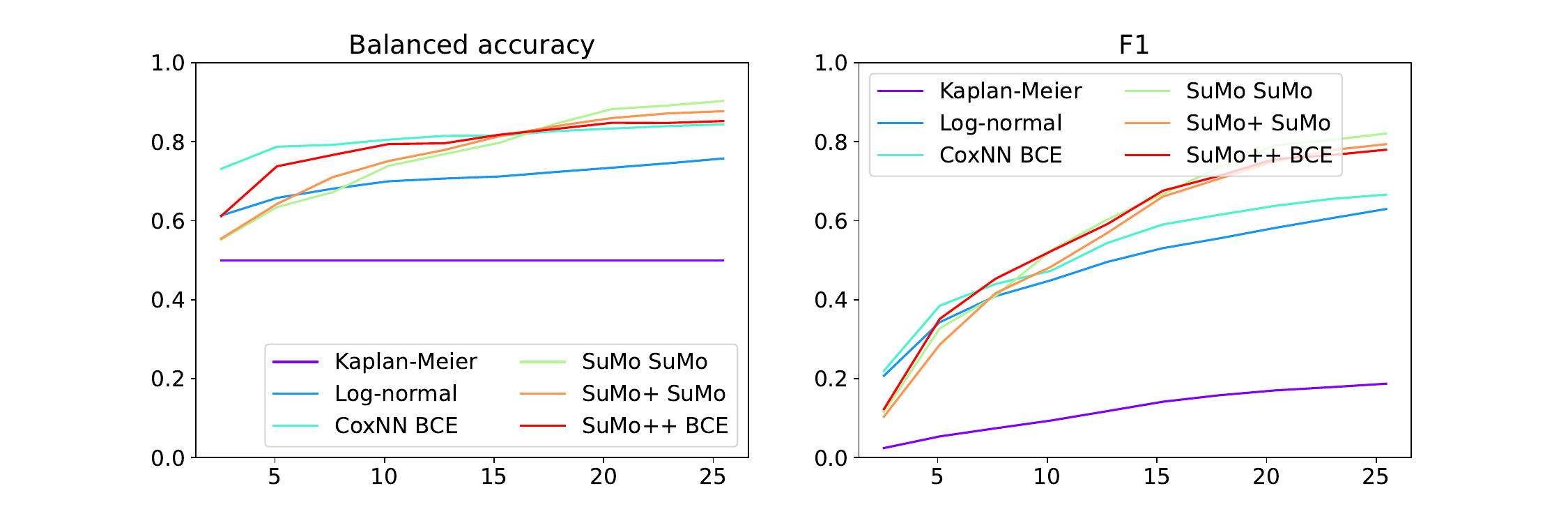}
    \caption{The balanced accuracy and the $F_1$ score over time for the COVID-19 dataset.}
    \label{fig:ot_covid}
\end{figure*}

\begin{figure*}
    \centering
    \includegraphics[trim={35mm 5mm 35mm 5mm}, clip, width=1\textwidth]{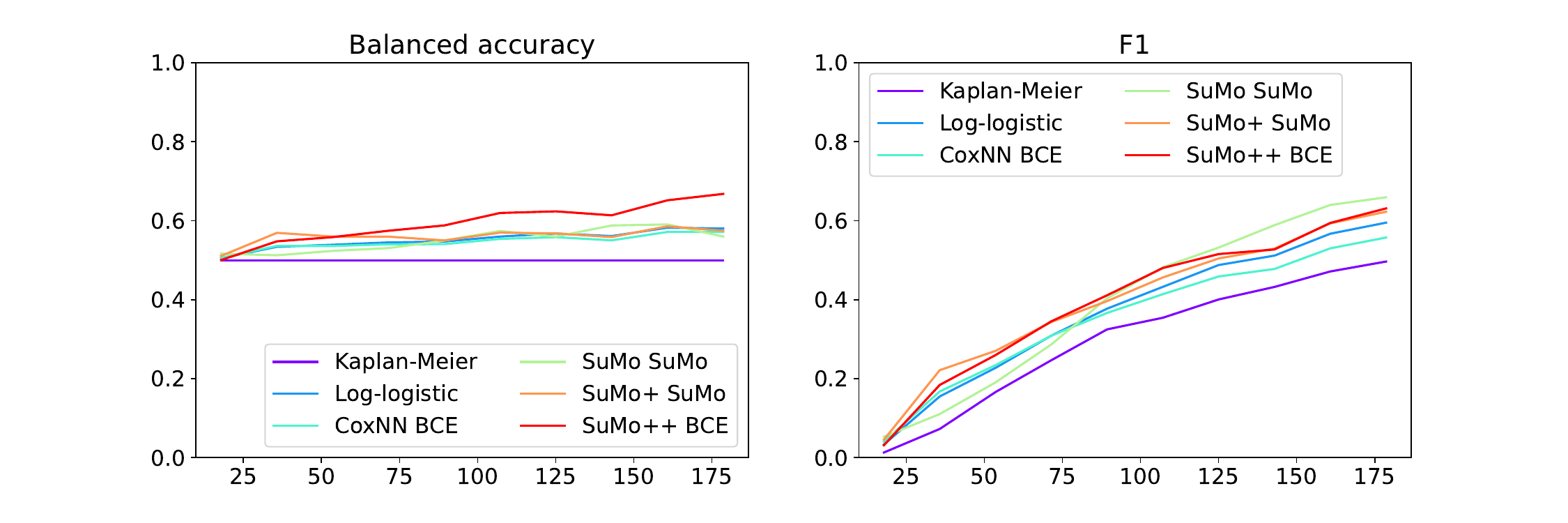}
    \caption{The balanced accuracy and the $F_1$ score over time for the GBSG2 dataset.}
    \label{fig:ot_gbsg2}
\end{figure*}

\begin{figure*}
    \centering
    \includegraphics[trim={35mm 5mm 35mm 5mm}, clip, width=1\textwidth]{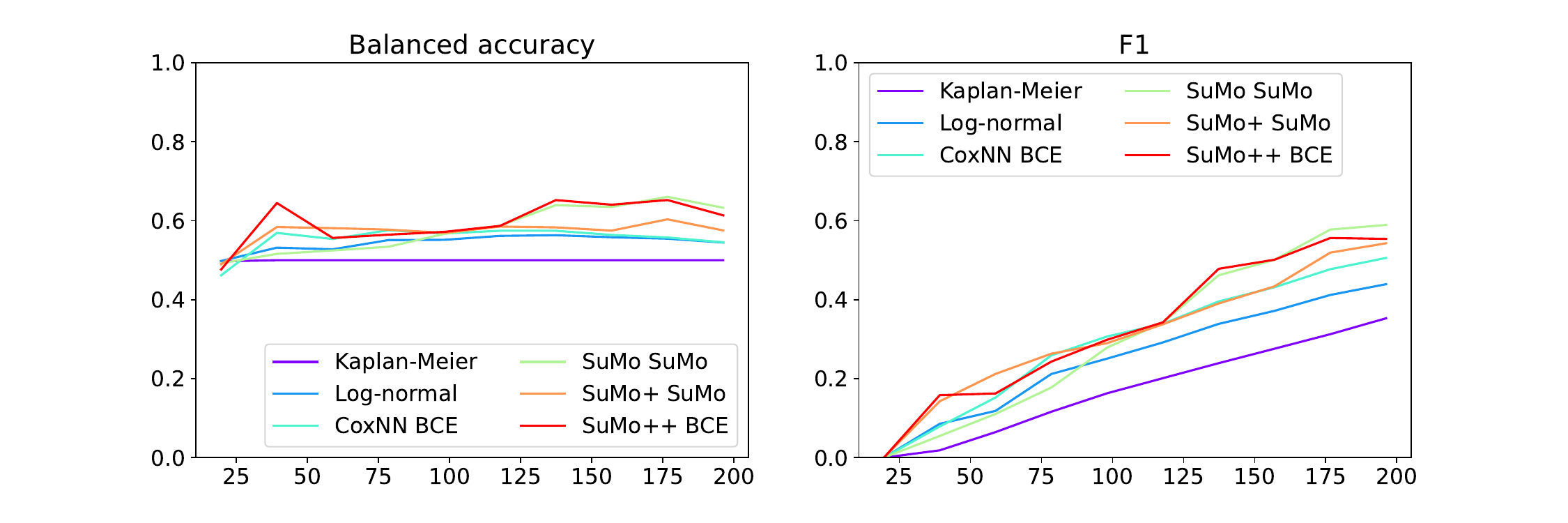}
    \caption{The balanced accuracy and the $F_1$ score over time for the Lymph dataset.}
    \label{fig:ot_lymph}
\end{figure*}

\begin{figure*}
    \centering
    \includegraphics[trim={35mm 5mm 35mm 5mm}, clip, width=1\textwidth]{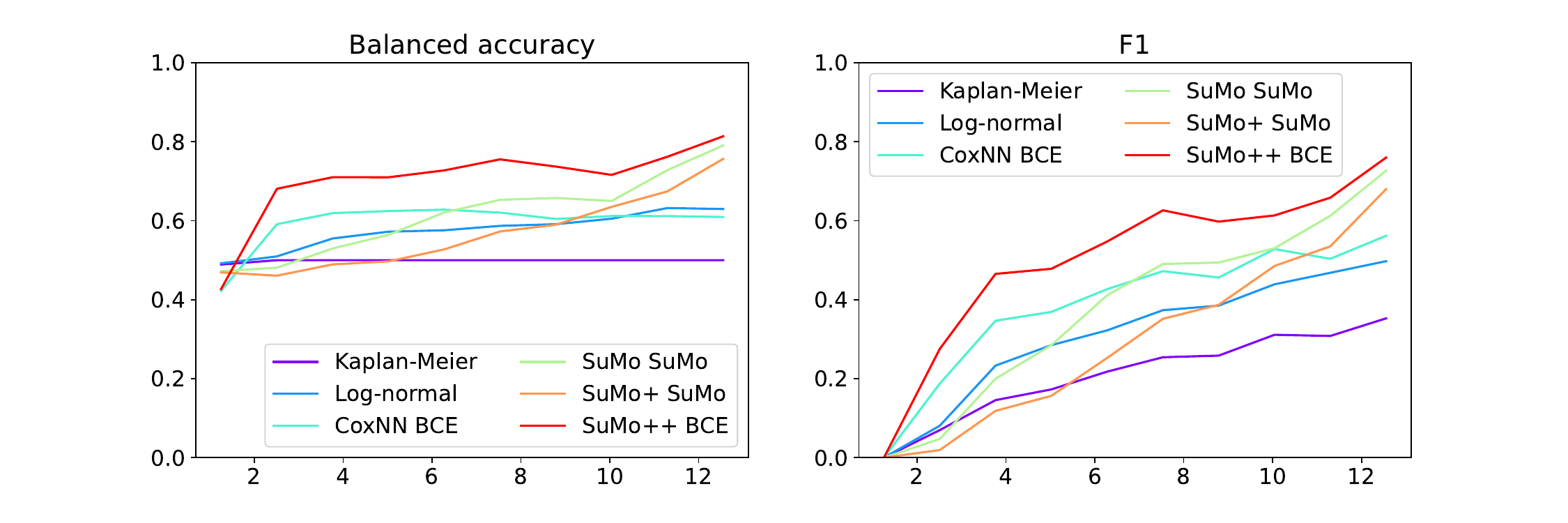}
    \caption{The balanced accuracy and the $F_1$ score over time for the NKI dataset.}
    \label{fig:ot_nki}
\end{figure*}

\begin{figure*}
    \centering
    \includegraphics[trim={35mm 5mm 35mm 5mm}, clip, width=1\textwidth]{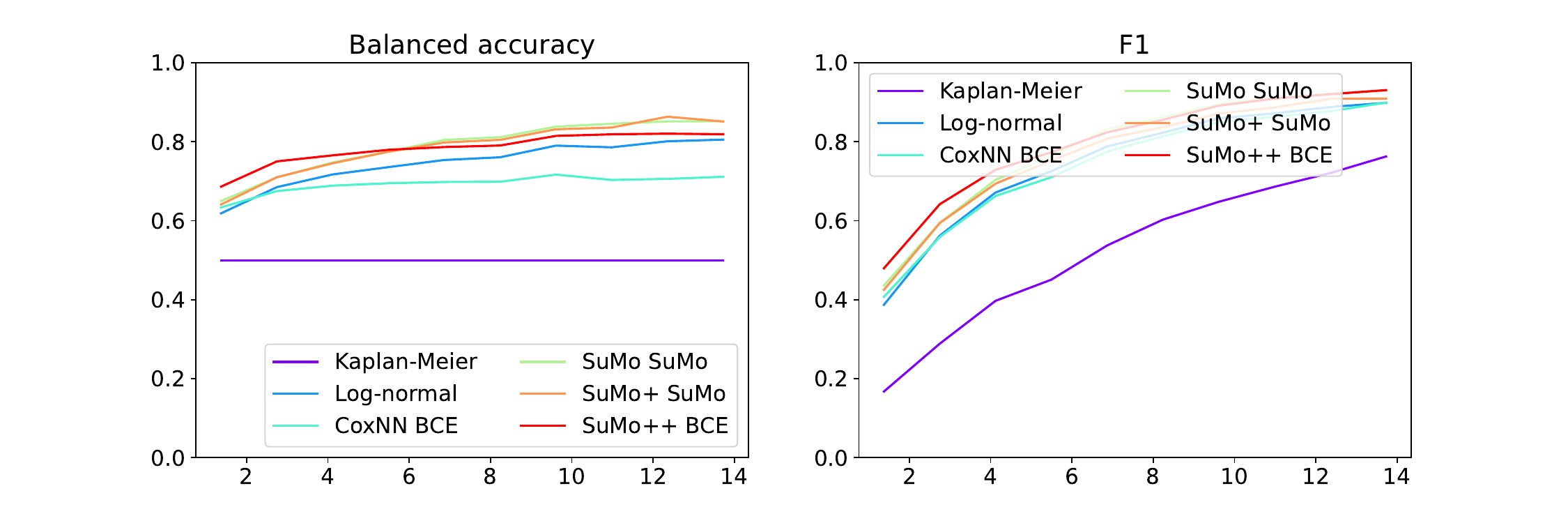}
    \caption{The balanced accuracy and the $F_1$ score over time for the Recur dataset.}
    \label{fig:ot_recur}
\end{figure*}

Here we provide the Figures~\ref{fig:ot_california},~\ref{fig:ot_clocks},~\ref{fig:ot_covid},~\ref{fig:ot_gbsg2},~\ref{fig:ot_lymph},~\ref{fig:ot_nki},~\ref{fig:ot_recur}. Each presents the non-integrated versions of the balanced accuracy and the $F_1$ score for the most prominent models -- one figure for each dataset. We chose these two scores as they are highly correlated with the mean of the scores and seem to complement each other.

\FloatBarrier
\section{Example plots of predicted survival curves}
\label{sec:predicted_survival_curves}
\begin{figure*}
    \centering
    \includegraphics[trim={5mm 5mm 5mm 5mm}, clip, width=1\textwidth]{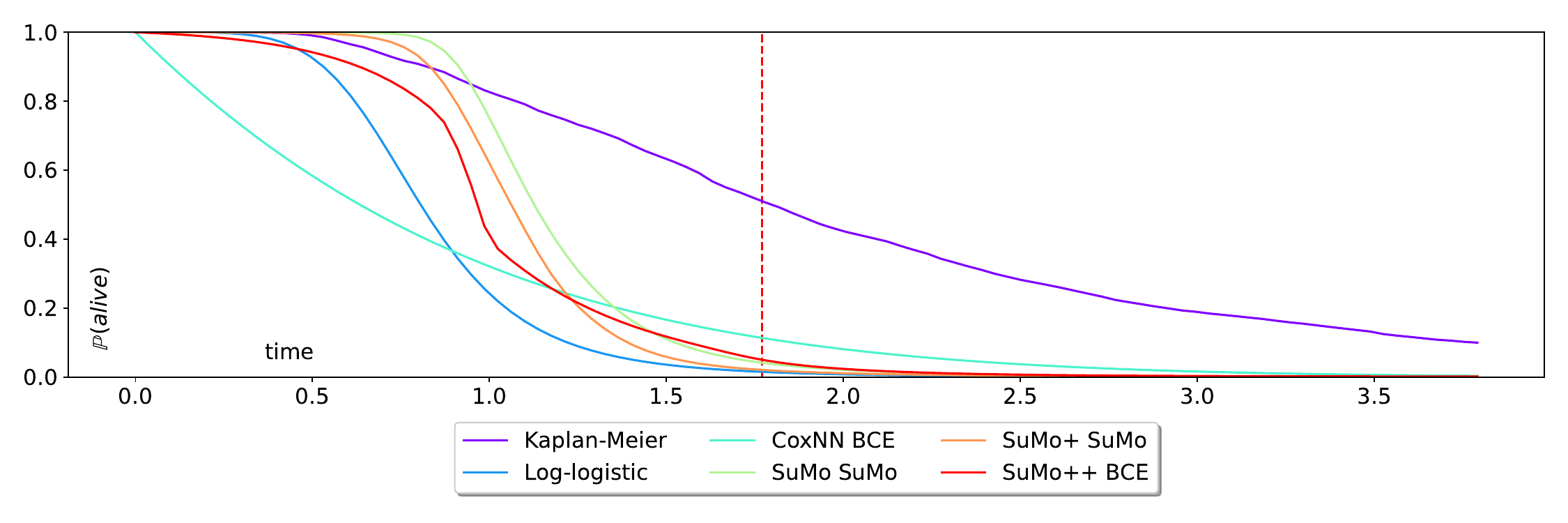}
    \caption{Survival curve examples for a sample from the California dataset up until $T_{max}$. The vertical line marks the time of the event or censoring (red: death, blue: censoring). The legend shows the model and, if applicable, the loss function (SuMo or BCE).}
    \label{fig:curve_example_0}
\end{figure*}

\begin{figure*}
    \centering
    \includegraphics[trim={5mm 5mm 5mm 5mm}, clip, width=1\textwidth]{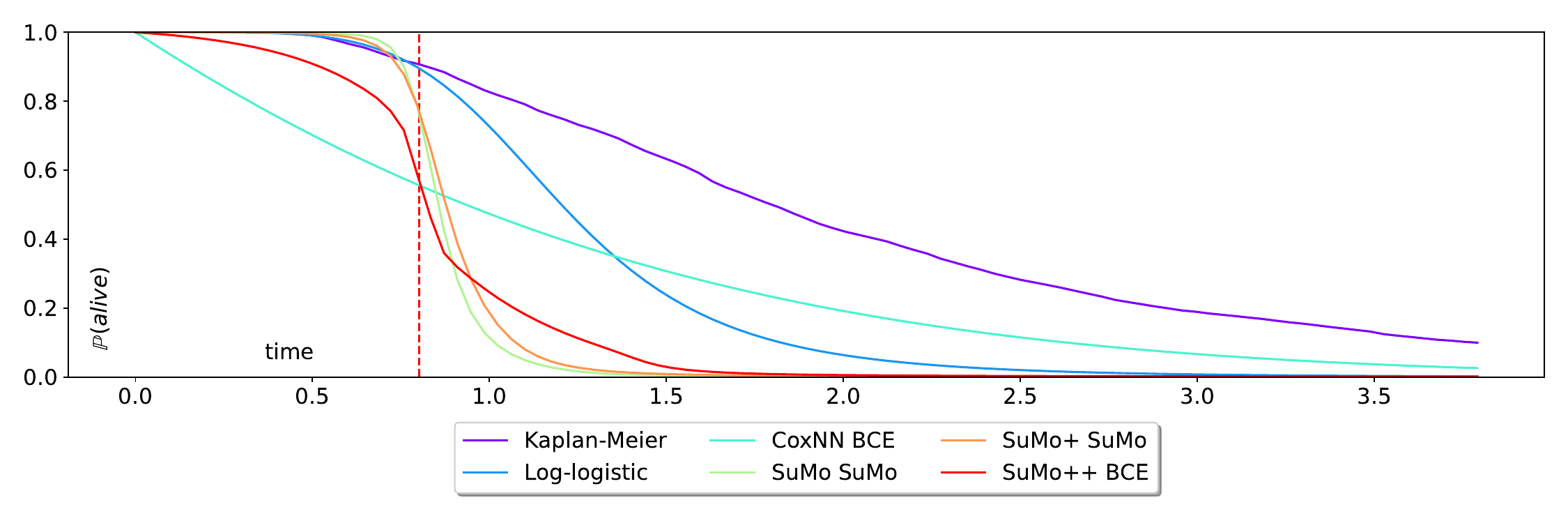}
    \caption{Survival curve examples for a sample from the California dataset up until $T_{max}$. The vertical line marks the time of the event or censoring (red: death, blue: censoring). The legend shows the model and, if applicable, the loss function (SuMo or BCE).}
    \label{fig:curve_example_1}
\end{figure*}

\begin{figure*}
    \centering
    \includegraphics[trim={5mm 5mm 5mm 5mm}, clip, width=1\textwidth]{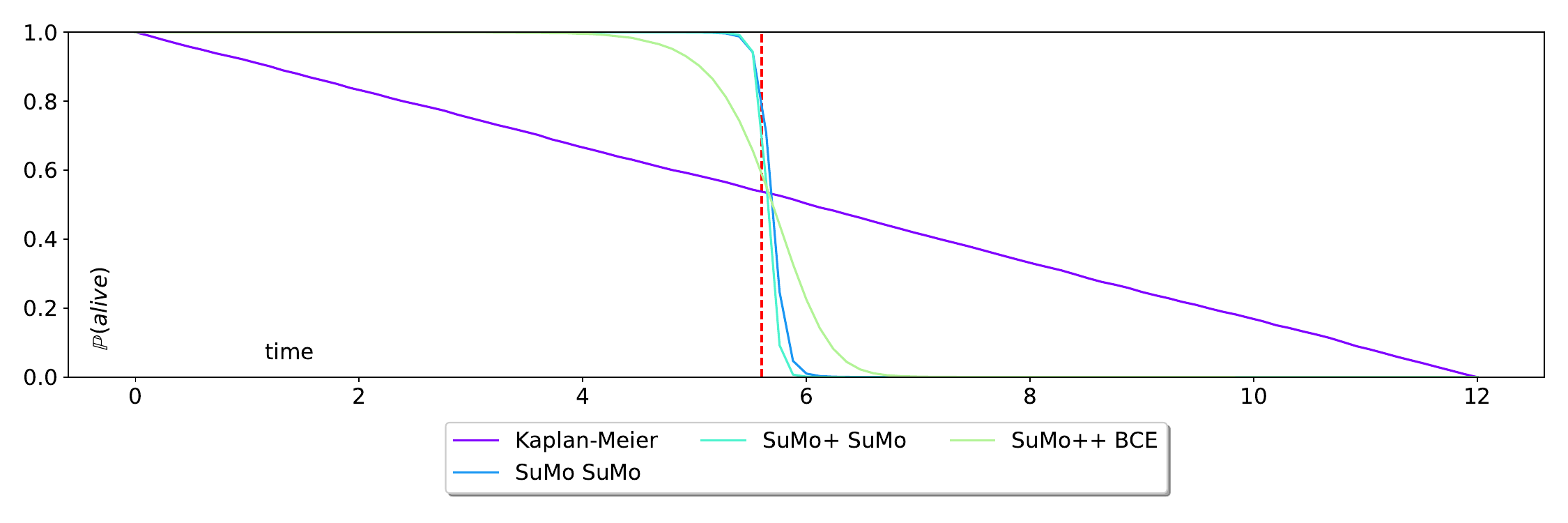}
    \caption{Survival curve examples for a sample from the Clocks dataset up until $T_{max}$. The vertical line marks the time of the event or censoring (red: death, blue: censoring). The legend shows the model and, if applicable, the loss function (SuMo or BCE).}
    \label{fig:curve_example_2}
\end{figure*}

\begin{figure*}
    \centering
    \includegraphics[trim={5mm 5mm 5mm 5mm}, clip, width=1\textwidth]{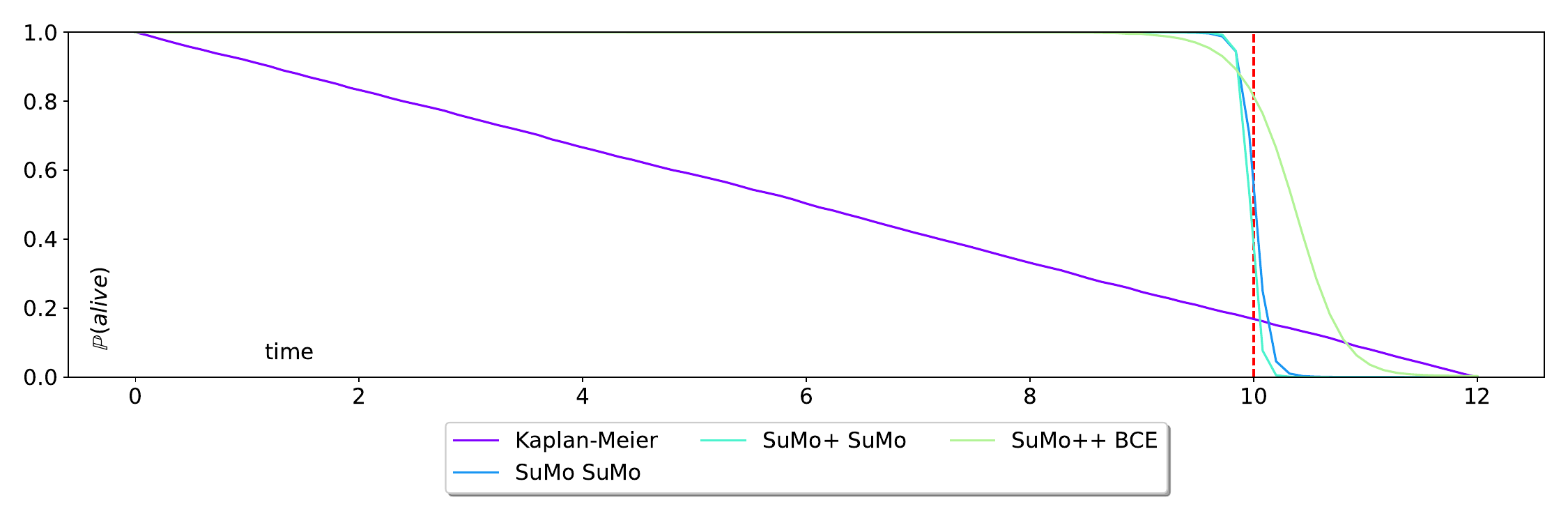}
    \caption{Survival curve examples for a sample from the Clocks dataset up until $T_{max}$. The vertical line marks the time of the event or censoring (red: death, blue: censoring). The legend shows the model and, if applicable, the loss function (SuMo or BCE).}
    \label{fig:curve_example_3}
\end{figure*}

\begin{figure*}
    \centering
    \includegraphics[trim={5mm 5mm 5mm 5mm}, clip, width=1\textwidth]{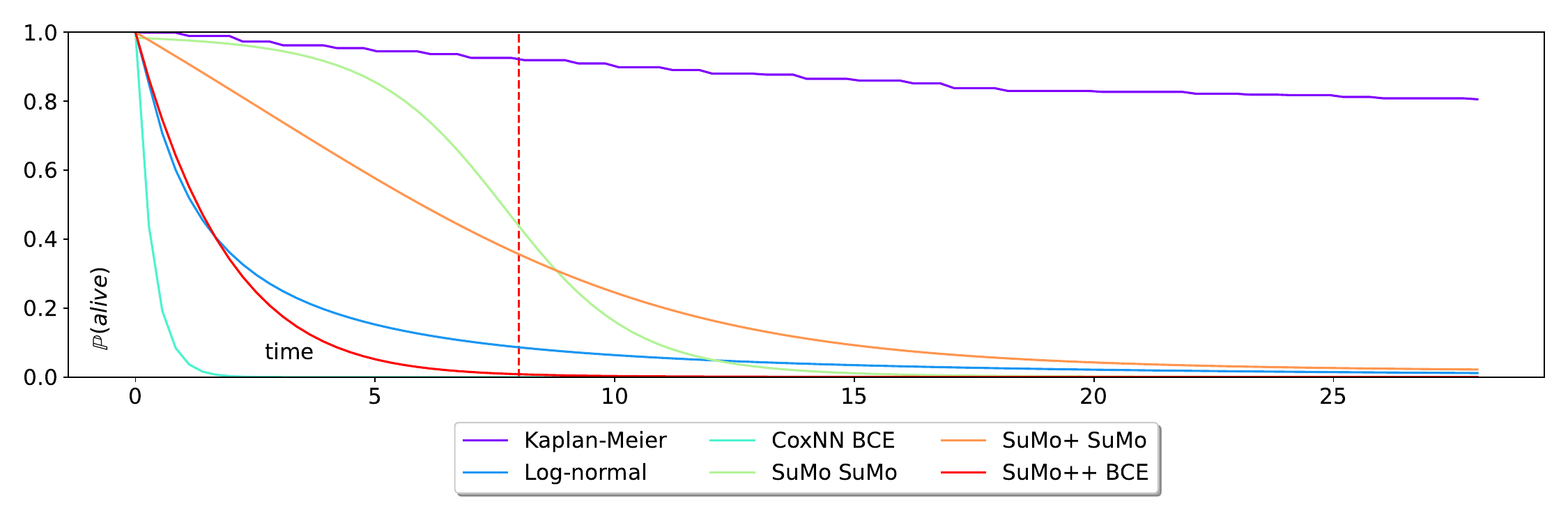}
    \caption{Survival curve examples for a sample from the COVID dataset up until $T_{max}$. The vertical line marks the time of the event or censoring (red: death, blue: censoring). The legend shows the model and, if applicable, the loss function (SuMo or BCE).}
    \label{fig:curve_example_4}
\end{figure*}

\begin{figure*}
    \centering
    \includegraphics[trim={5mm 5mm 5mm 5mm}, clip, width=1\textwidth]{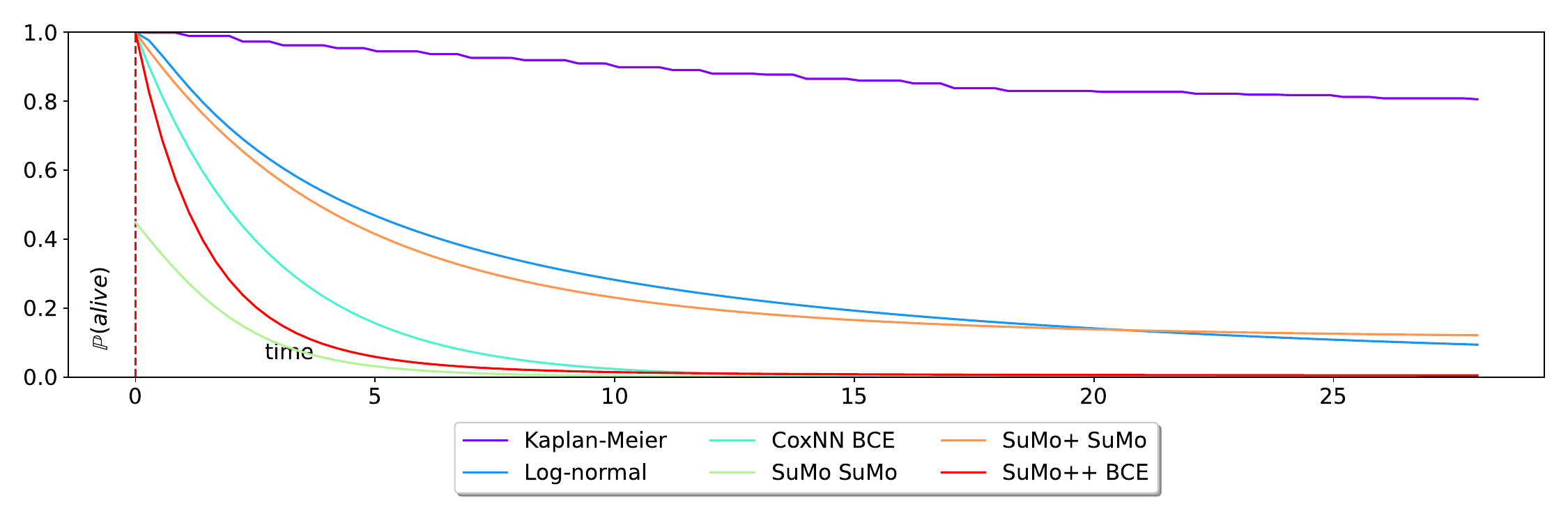}
    \caption{Survival curve examples for a sample from the COVID dataset up until $T_{max}$. The vertical line marks the time of the event or censoring (red: death, blue: censoring). The legend shows the model and, if applicable, the loss function (SuMo or BCE).}
    \label{fig:curve_example_5}
\end{figure*}

\begin{figure*}
    \centering
    \includegraphics[trim={5mm 5mm 5mm 5mm}, clip, width=1\textwidth]{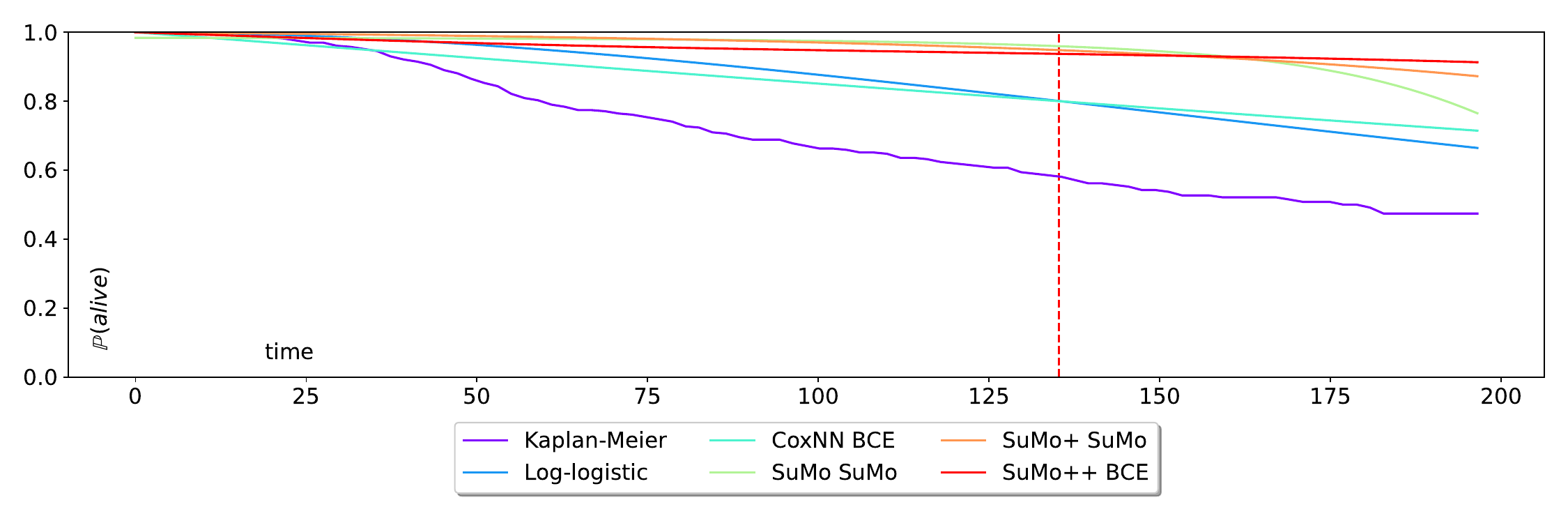}
    \caption{Survival curve examples for a sample from the GBSG2 dataset up until $T_{max}$. The vertical line marks the time of the event or censoring (red: death, blue: censoring). The legend shows the model and, if applicable, the loss function (SuMo or BCE).}
    \label{fig:curve_example_6}
\end{figure*}

\begin{figure*}
    \centering
    \includegraphics[trim={5mm 5mm 5mm 5mm}, clip, width=1\textwidth]{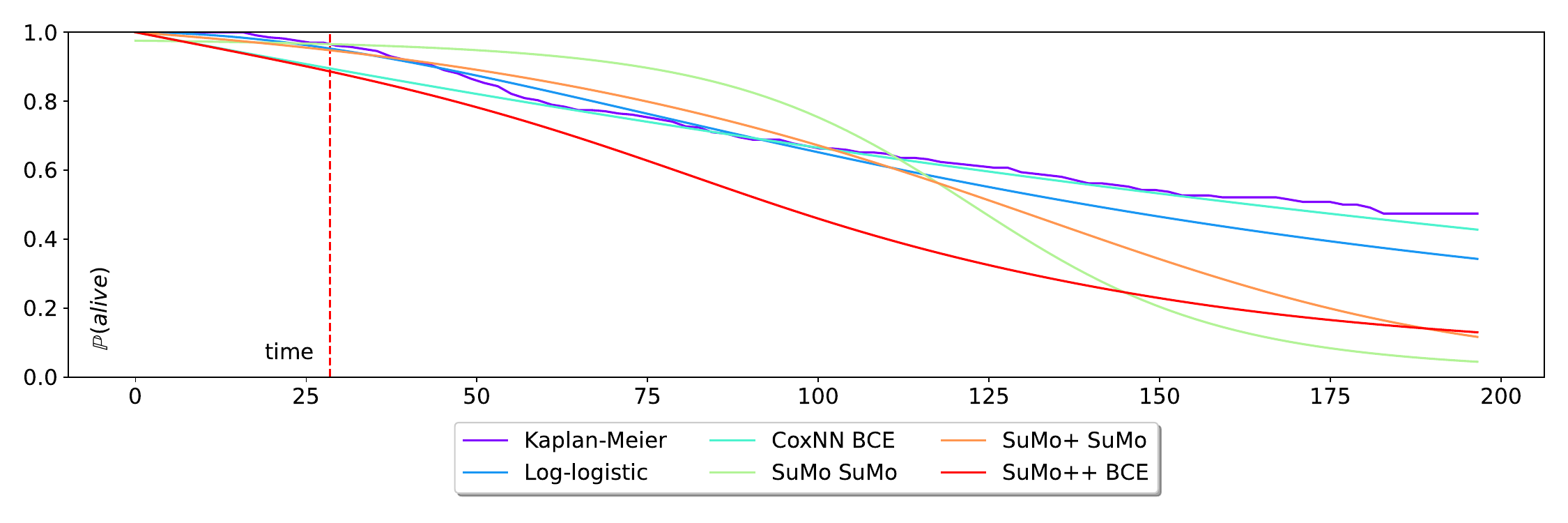}
    \caption{Survival curve examples for a sample from the GBSG2 dataset up until $T_{max}$. The vertical line marks the time of the event or censoring (red: death, blue: censoring). The legend shows the model and, if applicable, the loss function (SuMo or BCE).}
    \label{fig:curve_example_7}
\end{figure*}

\begin{figure*}
    \centering
    \includegraphics[trim={5mm 5mm 5mm 5mm}, clip, width=1\textwidth]{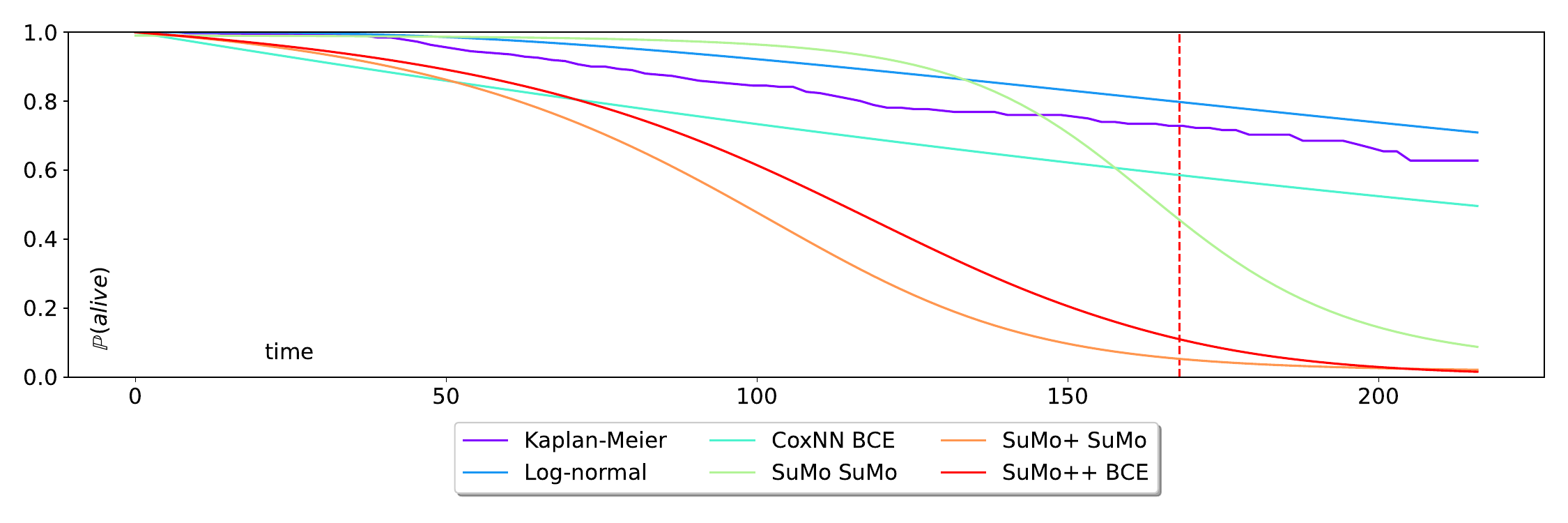}
    \caption{Survival curve examples for a sample from the Lymph dataset up until $T_{max}$. The vertical line marks the time of the event or censoring (red: death, blue: censoring). The legend shows the model and, if applicable, the loss function (SuMo or BCE).}
    \label{fig:curve_example_8}
\end{figure*}

\begin{figure*}
    \centering
    \includegraphics[trim={5mm 5mm 5mm 5mm}, clip, width=1\textwidth]{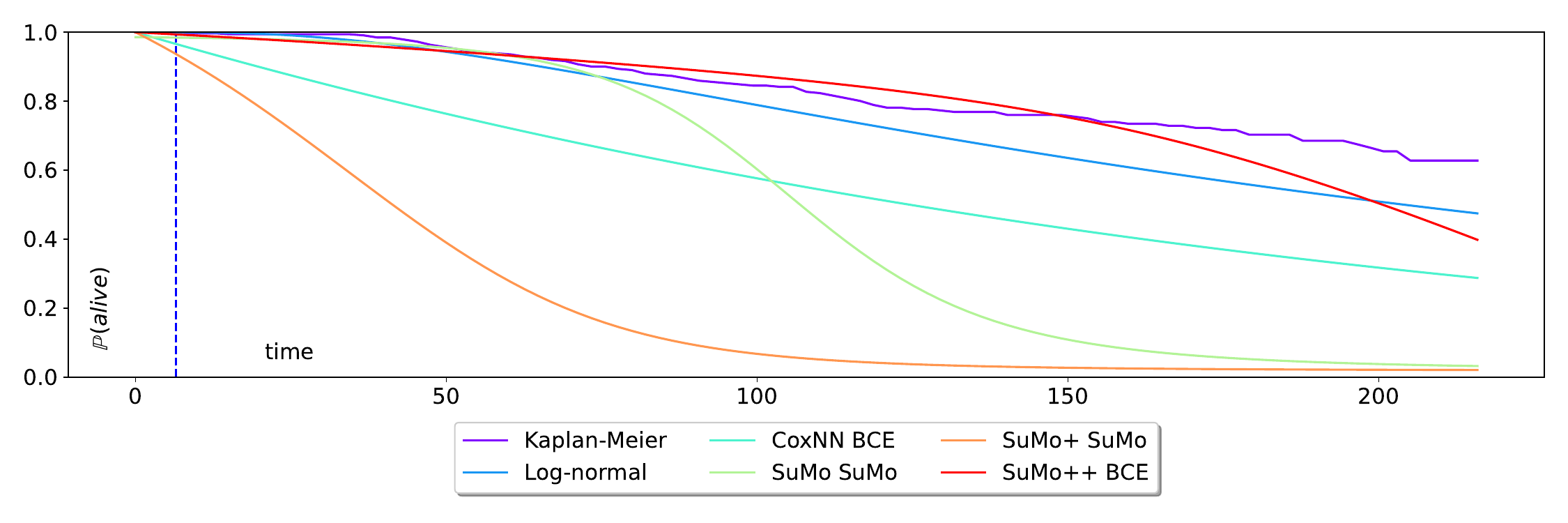}
    \caption{Survival curve examples for a sample from the Lymph dataset up until $T_{max}$. The vertical line marks the time of the event or censoring (red: death, blue: censoring). The legend shows the model and, if applicable, the loss function (SuMo or BCE).}
    \label{fig:curve_example_9}
\end{figure*}

\begin{figure*}
    \centering
    \includegraphics[trim={5mm 5mm 5mm 5mm}, clip, width=1\textwidth]{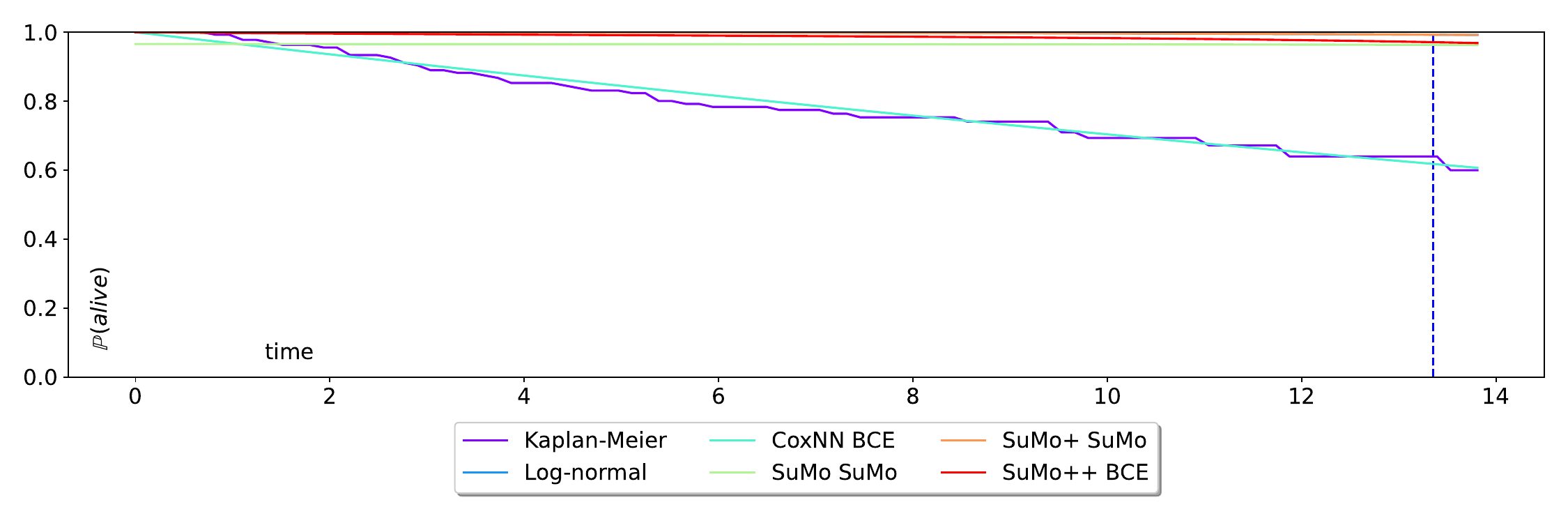}
    \caption{Survival curve examples for a sample from the NKI dataset up until $T_{max}$. The vertical line marks the time of the event or censoring (red: death, blue: censoring). The legend shows the model and, if applicable, the loss function (SuMo or BCE).}
    \label{fig:curve_example_10}
\end{figure*}

\begin{figure*}
    \centering
    \includegraphics[trim={5mm 5mm 5mm 5mm}, clip, width=1\textwidth]{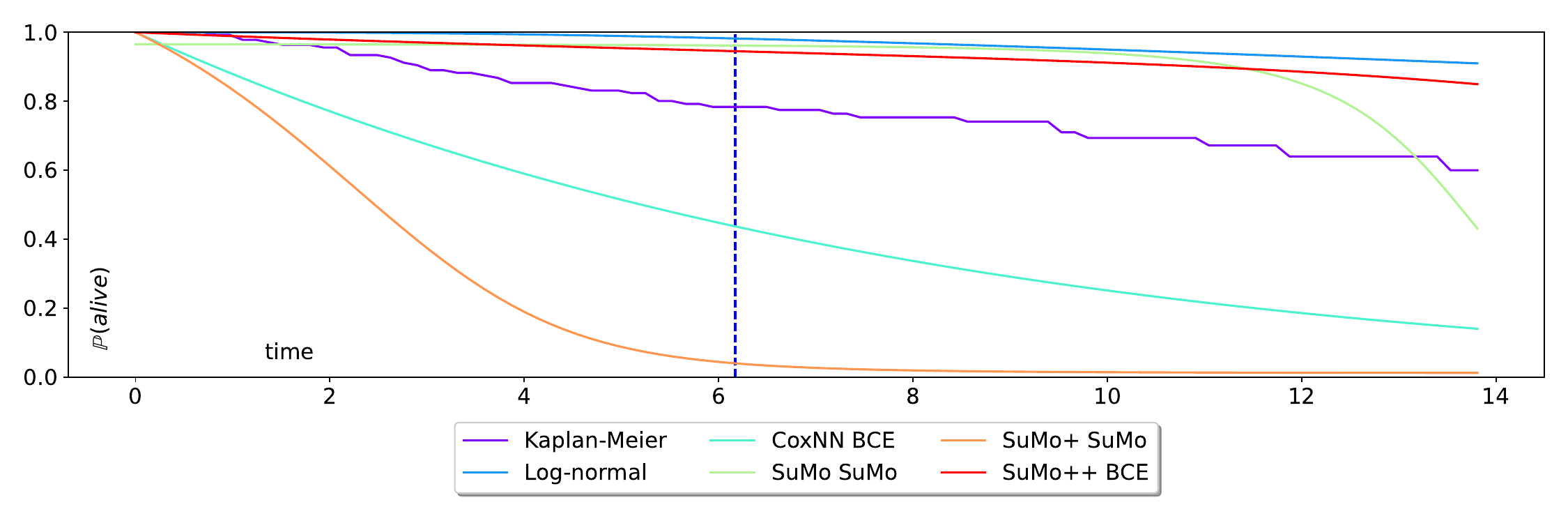}
    \caption{Survival curve examples for a sample from the NKI dataset up until $T_{max}$. The vertical line marks the time of the event or censoring (red: death, blue: censoring). The legend shows the model and, if applicable, the loss function (SuMo or BCE).}
    \label{fig:curve_example_11}
\end{figure*}

\begin{figure*}
    \centering
    \includegraphics[trim={5mm 5mm 5mm 5mm}, clip, width=1\textwidth]{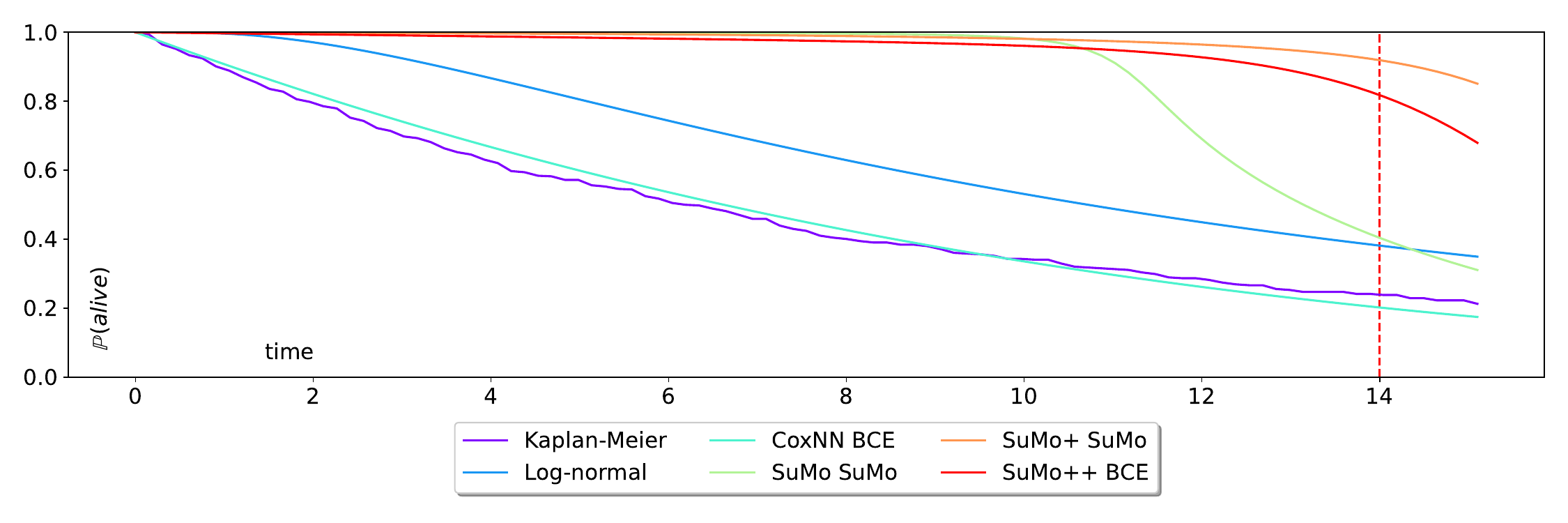}
    \caption{Survival curve examples for a sample from the Recur dataset up until $T_{max}$. The vertical line marks the time of the event or censoring (red: death, blue: censoring). The legend shows the model and, if applicable, the loss function (SuMo or BCE).}
    \label{fig:curve_example_12}
\end{figure*}

\begin{figure*}
    \centering
    \includegraphics[trim={5mm 5mm 5mm 5mm}, clip, width=1\textwidth]{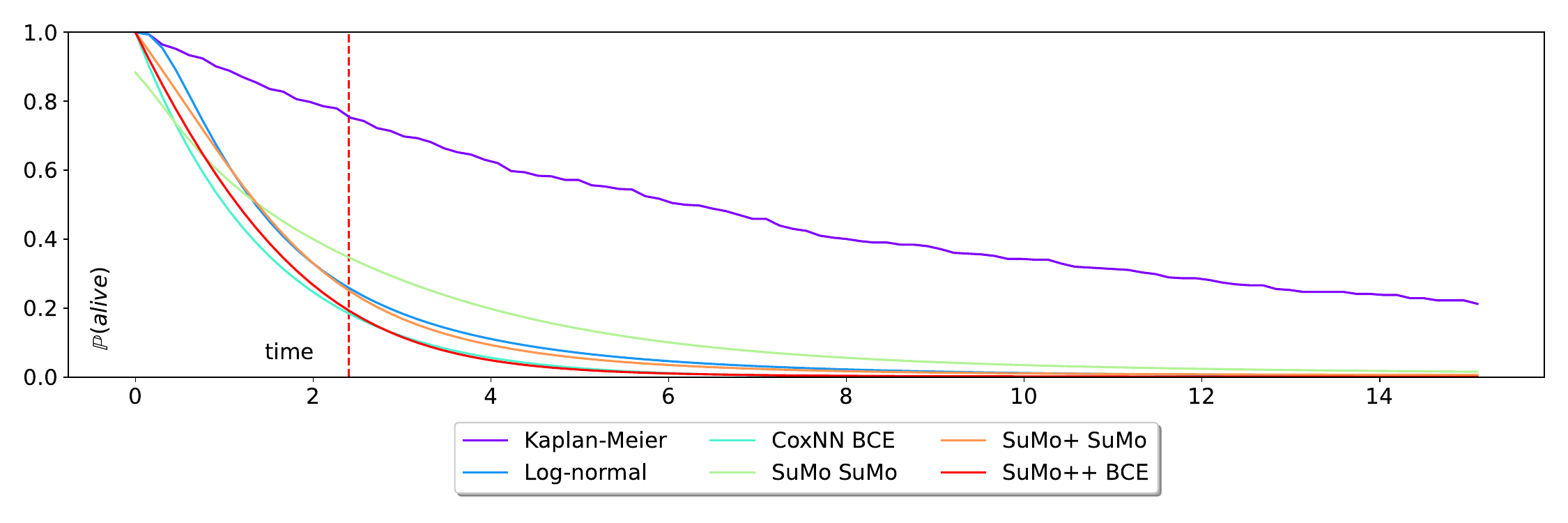}
    \caption{Survival curve examples for a sample from the Recur dataset up until $T_{max}$. The vertical line marks the time of the event or censoring (red: death, blue: censoring). The legend shows the model and, if applicable, the loss function (SuMo or BCE).}
    \label{fig:curve_example_13}
\end{figure*}

For the sake of readability, the Figures~\ref{fig:curve_example_0}, \ref{fig:curve_example_1}, \ref{fig:curve_example_2}, \ref{fig:curve_example_3}, \ref{fig:curve_example_4}, \ref{fig:curve_example_5}, \ref{fig:curve_example_6}, \ref{fig:curve_example_7}, \ref{fig:curve_example_8}, \ref{fig:curve_example_9}, \ref{fig:curve_example_10}, \ref{fig:curve_example_11}, \ref{fig:curve_example_12}, \ref{fig:curve_example_13} only show a selection of the trained models. For the classical models, we only select the one with the highest score in Table~\ref{tab:stats_Mean}.

\end{document}